\documentclass{clv3}

\usepackage{hyperref}
\usepackage{xcolor}
\usepackage{amsmath,amsfonts, amssymb,verbatim,ifthen}
\usepackage{amsthm}
\usepackage{color}
\usepackage{multirow}
\usepackage[all]{xy}
\usepackage{setspace}
\usepackage{graphicx}
\usepackage{url}
\usepackage{booktabs}
\usepackage{nameref}
\usepackage{latexsym}
\usepackage{csquotes}
\usepackage[leftmargin=3em]{quoting}
\usepackage{framed}
\usepackage{varwidth}
\usepackage{subcaption}
\usepackage{placeins}
\usepackage{lscape} 
\usepackage{enumitem}
\usepackage{bbm}
\usepackage{arydshln}

\DeclareMathOperator*{\argmax}{argmax}

\newcommand{\com}[1]{}

\newboolean{showcomments}
\setboolean{showcomments}{true}
\ifthenelse{\boolean{showcomments}}
  {\newcommand{\nb}[3]
    {
    {\color{#2}\small\fbox{\bfseries\sffamily\scriptsize#1}}
    {\color{#2}\sffamily\small$\triangleright~$\textit{\small #3}$~\triangleleft$}
    }
  }
  {    \newcommand{\nb}[3]{}  }

\definecolor{darkblue}{rgb}{0, 0, 0.5}
\hypersetup{colorlinks=true,citecolor=darkblue, linkcolor=darkblue, urlcolor=darkblue}

\runningtitle{Predicting Actions from Interviews}
\runningauthor{Oved*, Feder* and Reichart}

\begin{document}

\title{Predicting In-game Actions from Interviews of NBA Players}

\author{Nadav Oved \thanks{Authors contributed equally.} \\
  {\tt nadavo@campus.technion.ac.il} \\
  Amir Feder \footnotemark[1] \\
  {\tt feder@campus.technion.ac.il} \\
  Roi Reichart \\
  {\tt roiri@ie.technion.ac.il} \\}

\maketitle

\begin{abstract}
Sports competitions are widely researched in computer and social science, with the goal of understanding how players act under uncertainty. While there is an abundance of computational work on player metrics prediction based on past performance, very few attempts to incorporate out-of-game signals have been made. Specifically, it was previously unclear whether linguistic signals gathered from players' interviews can add information which does not appear in performance metrics. To bridge that gap, we define text classification tasks of predicting deviations from mean in NBA players' in-game actions, which are associated with strategic choices, player behavior and risk, using their choice of language prior to the game. We collected a dataset of transcripts from key NBA players' pre-game interviews and their in-game performance metrics, totalling in 5,226 interview-metric pairs. We design neural models for players' action prediction based on increasingly more complex aspects of the language signals in their open-ended interviews. Our models can make their predictions based on the textual signal alone, or on a combination of that signal with signals from past-performance metrics. Our text-based models outperform strong baselines trained on performance metrics only, demonstrating the importance of language usage for action prediction. Moreover, the models that employ both textual input and past-performance metrics produced the best results. Finally, as neural networks are notoriously difficult to interpret, we propose a method for gaining further insight into what our models have learned. Particularly, we present an LDA-based analysis, where we interpret model predictions in terms of correlated topics. We find that our best performing textual model is most associated with topics that are intuitively related to each prediction task and that better models yield higher correlation with more informative topics.\footnote{Code is available at: \url{https://github.com/nadavo/mood}}
\end{abstract}

\newpage
\section{Introduction}
\label{sec:intro}

Decision theory is a well-studied field, with a variety of contributions in economics, statistics, biology, psychology and computer science~\cite{berger1985statistical, einhorn1981behavioral}. While substantial progress has been made in analyzing the choices agents make, prediction in decision making is not as commonly researched, partly due to its challenging nature~\cite{gilboa2009theory}. Particularly, defining and assessing the set of choices in a real-world scenario is difficult, as the full set of options an agent faces is usually unobserved, and her decisions are only inferred from their outcomes. 

One domain where the study of human action is well defined and observable is sports, in our case Basketball. Professional athletes are experts in decision making under uncertainty, and their actions, along with their outcomes, are well-documented and extensively studied. While there are many attempts to predict game outcomes in Basketball, including win probability, players' marginal effects and the strengths of specific lineups~\cite{gangulyproblem, coate2012basic}, they are less focused on the decisions of individual players.

Individual player actions are difficult to predict as they are not made in lab conditions and are also a function of "soft" factors such as their subjective feelings regarding their opponents, teammates and themselves. Moreover, such actions are often made in response to the decisions their opponents and teammates make. Currently, sports analysts and statisticians that try to predict such actions do so mostly through past performances, and their models do not account for factors such as those mentioned above~\cite{kaya2014decision}.

However, there is an additional signal, ingrained in fans' demand for understanding the players current state - pre-game interviews. In widely successful sports such as baseball, football and basketball, top players and coaches are regularly interviewed before and after games. These interviews are usually conducted to get a glimpse of how they are currently feeling and allow them to share their thoughts, given the specifics of the upcoming game and the baggage they are carrying from previous games. Following the sports psychology literature, we wish to employ these interviews to gain an insight about the players' emotional state and its relation to actions~\cite{uphill2014influence}.\footnote{Building on this literature, we use this concept of "emotional state" freely here and note that while some similarities exist, it is not directly mapped to the psychological literature.}

In the sports psychology literature, there is a long standing attempt to map the relationship between what this literature defines as "emotional state" and performance. The most popular account of such a relationship is the model of Individual Zones of Optimal Functioning (IZOF)~\cite{hanin1997emotions}. IZOF proposes that there are individual differences in the way athletes react to their emotional state, with each having an optimal level of intensity for each emotion for achieving top performance. IZOF suggests viewing emotions from a utilitarian perspective, looking at their helpfulness in achieving individual and team goals, and aims to calibrate the optimal emotional state for each player to perform at her best.

In this paper we build on that literature and aim to predict actions in Basketball, using the added signal provided in the interviews. We explore a multi-modal learning scheme, exploiting player interviews alongside performance metrics or without them. We build models that use as input the text alone, the metrics, and both modalities combined. As we wish to test for the predictive power of language, alone or in combination with past performance metrics, we look at all 3 settings, and discuss the learned representation of the text modality with respect to the "emotional state" that could be captured through the model.

We treat the player's deviation from his mean performance measure in recent games as an indication to the actions made in the current game. By learning a mapping from players' answers to underlying performance changes we hope to integrate a signal about their thoughts into the action prediction process. Our choice to focus on deviations from the mean performance and not on absolute performance values, is also useful from a machine learning perspective: It allows us to generalize across players, despite the differences in their absolute performance. We leave a more in-depth discussion of the formulation of our prediction task for later in the paper (Section \ref{sec:task}).

Being interested in the added behavioral signal hidden in the text, we focus on the task (Section~\ref{sec:task}) of predicting metrics that are associated in the literature with in-game behavior and are endogenous to the player's strategic choices and mental state: shot success share on the offensive side, and fouls on the defensive side~\cite{goldman2011allocative}. We further add our own related metrics: the player's mean shot location, his assists to turnovers ratio, and his share of 2 point to 3 point shot attempts. We choose to add those metrics as they are measurable on a play-by-play basis, and are interesting measures of relative risk. We believe our proposed measures can isolate to some extent the risk associated with specific types of decisions, such as when to pass, when to shoot and where to do it.

Almost no single play result is a function of only one player's action; Yet, our positive results from models that exploit signals from individual players only (Section~\ref{sec:results}) indicate that meaningful predictions can be made even without direct modeling of inter-player interactions. As this is a first paper on the topic, we leave for future work an exploration of how player interactions can be learned, noting that such an attempt will surely entail a more complex model.\footnote{There are novel attempts to estimate players' partial effect on the game~\cite{gramacy2017hockey}, which comprises of estimating the difference they make on final game outcomes. However, in this research we decided to focus on metrics that can be attributed to specific types of decisions and not to overall game outcomes.} Also, we believe that if the interviews provide a strong signal regarding players' in-game decisions, it should be observed even when interactions are not explored. Hopefully, our work will encourage future research that considers interactions as well.

We collected (Section~\ref{sec:data}) a dataset of 1,337 interviews with 36 major NBA players during a total of 14 seasons. Each interview is augmented with performance measures of the player in each period (quarter) of the corresponding game.
To facilitate learnability, we focus on NBA all-stars as they are consistently interviewed before games, and have played key roles throughout their career. Also, the fact that many players in our dataset are still active and are expected to remain so in the following years, gives us an opportunity to measure our model's performance and improve it in the future.

We start by looking at a regression model as a baseline for both the text-only and the metrics-only schemes. Then, we experiment with structure-aware neural networks for their feature learning capabilities and propose (Section~\ref{sec:models}) models based on LSTM \cite{hochreiter1997long} and CNN \cite{lecun1998gradient}. Finally, to better model the interview structure and to take advantage of recent advancements in contextual embeddings, we also use a BERT-based architecture \cite{vaswani2017attention, devlin2019bert} and explore the trade-off between a light-weight attention mechanism and more parameter heavy alternatives (Section~\ref{sec:models}).

Our results (Sections \ref{sec:experiments}, \ref{sec:results}) suggest that our text-based models are able to learn from the interviews to predict the player's performance metrics, while the performance-based baselines are not able to predict much better than a coin flip or the most common class, a phenomenon we try to explain in Section~\ref{sec:results}. 

Interestingly, the models that exploit both the textual signal and the signal from past performance metrics improve on some of the most challenging predictions. These results are consistent with the hypothesized relationship of mental state and performance, and support claims in the literature that such an "emotional state" has predictive power on player performance~\cite{lazarus2000emotions, hanin2007emotions}.

Our contributions to the sports analytics and NLP literature are as follows: (1) We provide the first model, as far as we know, that predicts player actions from language; (2) Our model is the first that can predict relative player performance without relying on past performance; (3) On a more conceptual level, our results suggest that the player's "emotional state" is related to player performance.

We support our findings with a newly-proposed approach to qualitative analysis (Section~\ref{sec:qual}). As neural networks are notoriously difficult to directly interpret, we choose to analyse text-based NN models via topic modeling of the texts associated with the model's predictions. Alternative approaches for model explanations do not allow for reasoning over higher-level concepts such as topics. Hence, we believe this could be a beneficial way to examine many neural models in NLP and view this as an additional contribution we present in this paper. 

This analysis includes a comparison of our best performing models and finds that our BERT-based model is most associated with topics that are intuitively related to each prediction task, suggesting that the hypothesized "emotional state" from the sports psychology literature could have been learned. Additionally, we find that this correlation becomes stronger as the confidence of the model in its prediction increases, meaning that a higher probability for such topics corresponds to higher model confidence. Finally, we compare our models and observe that better performing models yield higher correlation with more informative topics.

In conclusion, we believe that this paper provides evidence for the transmission of language into human actions. We demonstrate that our models are able to predict real world variables via text, extending a rich NLP tradition and literature about tasks such as sentiment analysis, stance classification and intent detection that also extract information regarding the text author. We hope this research problem and the high level topic will be of interest to the NLP community. To facilitate further research we also release our data and code.
\section{Related Work}
\label{sec:related}

Previous work on the intersection of language, behavior and sports is limited due to the rarity of relevant textual data~\cite{xu2015hidden}. However, there is an abundance of research on predicting human decision making (e.g.~\cite{rosenfeld2018predicting, plonsky2017psychological, hartford2016deep}), on using language to predict human behavior~\cite{sim2016friends, niculae2015linguistic} and on predicting outcomes in Basketball~\cite{gangulyproblem, cervone2014pointwise}. Since we aim to bridge the gap between the different disciplines, we survey the relevant work in each.

\subsection{Prediction and Decision Making}

Previous decision making work is both theoretical -- modelling the incentives  individuals face and the equilibrium observed given their competing interests~\cite{gilboa2009theory}, and empirical -- aiming to disentangle causal relationships that can shed light on what could be driving actions observed in the world~\cite{angrist2008mostly, kahneman1979prospect}. 

While there are some interesting attempts at learning to better predict human action~\cite{hartford2016deep, wright2010beyond, erev1998predicting}, the task at hand is usually addressed in lab conditions or using synthetic data. In a noisy environment it becomes much harder to define the choice set, that is the alternatives the agent faces, and to observe a clear outcome, the result of the action taken. In our setting we can only observe proxies to the choices made, and they can only be measured discretely, whenever a play is complete. Moreover, we can not easily disentangle the outcome of the play from the choices that drove it, since actions are dependent on both teammates and adversaries. 

Our work attempts to integrate linguistics signals into a decision prediction process. Language usage seems to be informative about the speaker's current state of mind~\cite{wardhaugh2011introduction} and his personality~\cite{fasold1990sociolinguistics}. Yet, this is rarely explored in the context of decision making~\cite{gilboa2009theory}. Here we examine whether textual traces can facilitate predictions in decision making.

\subsection{NLP and Prediction of Human Behavior}

NLP algorithms, and particularly Deep Neural Networks (DNNs), often learn a  low dimensional language representation with respect to a certain objective and in a manner which preserves valuable information regarding the text or the agent producing it. For example, in sentiment analysis~\cite{pang2002thumbs} text written by different authors is analyzed with respect to the same objective of determining whether the text conveys positive or negative sentiment. This not only reveals something about the text, but also about the author -- her personal stance regarding the subject she was writing about. One can view our task to share some similarity with sentiment classification as both tasks aim to learn something about the emotional state of the author of a given text.

Yet, a key difference between the two tasks, which poses a greater challenge in our case, is that in our task the signal we are aiming to capture is not clearly visible in the text, and requires inferring more subtle or abstract concepts than positive or negative sentiment. Given a movie review, an observer can guess if it is positive or negative rather easily. In our case, it is unclear where in the text is the clue regarding the players' mental state, and it is even less clear how it will correspond to their actions. Moreover, the text in our task involves a form of structured dialog between two speakers (the player and the interviewer), which entails an additional level of complexity, on top of the internal structures present for each speaker independently.

In a sense, our question is actually broader. We want to examine whether textual traces can help us in the challenging problem of predicting human action. There is a long standing claim in the social sciences that one could learn information about a person's character and his behavior from their choice of language~\cite{fasold1990sociolinguistics,wardhaugh2011introduction,bickerton1995language}, but this claim was not put to test in a real world setting such as those we are testing in. Granted, understanding character from language and predicting actions from language are quite different. However, if it is the case that neural networks could learn a character-like context using the final action as the supervision sign, it could have substantial implications for language processing and even the social sciences.

In the emerging field of computational social science, there is a substantial effort to harness linguistic signals to better answer scientific questions~\cite{danescu2013computational}. This approach, a.k.a text-as-data, has led to many advancements in the prediction of stock prices~\cite{kogan2009predicting}, understanding of political discourse~\cite{field2018framing} and analysis of court decisions~\cite{goldwasser2014object, sim2016friends}.
Our work adds another facet to this literature, trying to identify textual signals that enable the prediction of actions which are not explicitly mentioned in the text.

\subsection{Prediction and Analysis in Basketball}

Basketball is at the forefront of sports analytics.  In recent decades, there have been immense efforts to document every aspect of the game in real-time, and currently for every game there is data capturing each play's result, player and ball movements and even crowd generated noise.
Researchers have employed this data to solve  prediction tasks about game outcomes~\cite{gangulyproblem, kvam2006logistic}, points and performance~\cite{cervone2014pointwise, sampaio2015exploring}, and possession outcome~\cite{cervone2016multiresolution}.

Recent work has also explored mechanisms that facilitate the analysis of the decisions players and coaches make in a given match~\cite{kaya2014decision, bar2000criticality}. Some have tried to analyze the efficiency and optimality of decisions across the game~\cite{goldman2011allocative, wang2018advantage}, while others have focused on the decisions made in the final minutes of the game, when they are most critical~\cite{mcfarlaneevaluating}. Also, attempts were made to model strategic in-game interactions in order to simulate and analyze counterfactual scenarios~\cite{sandholtzreplaying} and to understand the interplay in dynamic space creation between offense and defense~\cite{lamas2015modeling}. We complement this literature by making text-based decision-related predictions. We address the player's behavior and current mental state as a factor in analyzing his actions, while previous work in sports analytics focused only on optimality considerations. Following the terminology of the sports psychology literature, we attempt to link players emotional/mental state, as manifested in the interviews, to the performance, actions and risk taking in the game~\cite{hanin1997emotions,uphill2014influence}.

\section{Data}
\label{sec:data}

We created our dataset with the requirement that we have enough data on both actions and language, from as many NBA seasons as possible and for a variety of players. While the number of seasons is constrained by the availability of transcribed interviews, we had some flexibility in choosing the players. To be able to measure a variety of actions and the corresponding interviews across time, we chose to focus on players that were important enough to be interviewed repeatedly and crucial enough for their team so that they play throughout most of the game. These choices allowed us to measure player performance not only at the game level, but also in shorter increments, such as the period level.

Our dataset is therefore a combination of two resources: (1) A publicly available play-by-play dataset, collected from \url{basketball-reference.com}; and (2) The publicly available interviews at \url{ASAPsports.com}, collected only for players that were interviewed in more than three different seasons. Interviews were gathered from the $2004-2005$ basketball season up until June $2018$. As this dataset comes from a fairly unexplored domain with regards to NLP, we follow here with a basic description of the different sources.
For a more detailed description and advanced statistics such as common topics, interview length and player performance distributions, please see Tables~\ref{tab:metrics},~\ref{tab:interviewsperplayer} and~\ref{tab:topics}.

We processed the play-by-play data to extract individual metrics for each player in each game for which that player was interviewed. The metrics were collected at both the game and the period level (see Section~\ref{sec:task} for the description of the metrics). We aggregated the performance metrics at the period level, to capture performance at different parts of the game and reduce the effects of outliers.\footnote{There are $4$ periods in a Basketball game, not including overtime. We do not deal with overtime performance as it might be less affected by the player pre-game state, and more by the happenings in the $4$ game periods.} This is important since performance in the first quarter could have a different meaning than performance in the last, where every mistake could be irreversible. Each interview consists of question-answer pairs for one specific player, and hence properties like the interview length and the length of the different answers are player specific. Key players are interviewed before each game, but we have data mostly for playoff games, since they were the ones that were transcribed and uploaded.\footnote{See Subsection~\ref{subsec:interviews} for an explanation on NBA playoffs.} This bias makes sense since  playoff interviews are more in-depth and they attract a larger audience. Overall, our dataset consists of $2,144$ interviews, with some players interviewed twice between consecutive games. After concatenating such interviews we are left with $1,337$ interviews from $36$ different players, and the corresponding game metrics for each interview. The total number of interview-period metric pairs is $5,226$.

We next describe our in-game play-by-play data and the pre-game interviews, along with the processing steps we apply to each.

\subsection{In-game Play-by-Play}

\begin{figure}
\centering
\includegraphics[width=\textwidth]{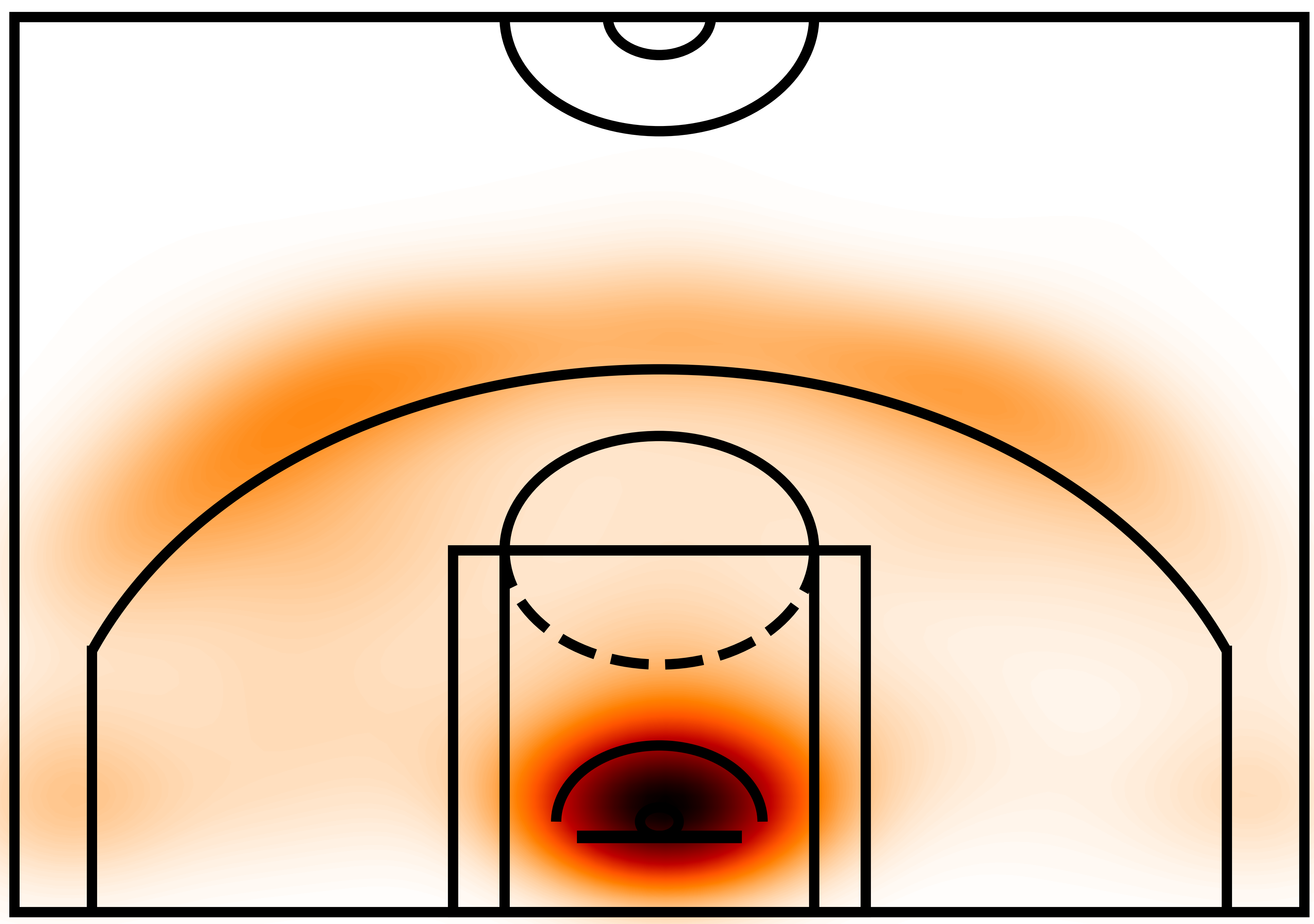}
\caption{Shot location of all attempted shots for all the players in our dataset. A darker color represents more shots attempted at that location. Black lines represent the structure of one of the two symmetric halves of an NBA basketball court.}
\label{fig:shot_loc}
\end{figure}

Basketball data is gathered after each play is done. As described in our "basketball dictionary" in Table~\ref{tab:bball_dict}, a play is any of the following events: Shot, Assist, Block, Miss, Free Throw, Rebound, Foul, Turnover, Violation, Time-out, Substitution, Jump Ball and Start and End of Period. We ignore Time-outs, Jump Balls, Substitutions and Start/End of Period plays as they do not add any information with respect to the metrics that we are monitoring. If a shot was successful, there could be an assist attributed to the passing player. Also, we observe for every foul the affected player and the opponent charged, as well as the player responsible for each shot, miss, free-throw and lost ball. For every shot taken, there are two location variables, indicating the shot's coordinates on the court, with which we calculate relative distance from the basket (see Figure~\ref{fig:shot_loc}). We use those indicators to produce performance metrics for each period.

\begin{table}[htbp]
    \centering
    \begin{tabular}{|p{2cm}|p{10cm}|} \hline
    Term & Description \\ \hline
    Shot  & Attempting to score points by throwing the ball through the basket. Each successful shot is worth $3$ points if behind the $3$ point arc, and $2$ otherwise. \\ \hline
    Assist & Passing the ball to a teammate that eventually scores without first passing the ball to any other player. \\ \hline
    Block & Altering an attempted shot by touching the ball while still in the air. \\ \hline
    Free Throw & Unopposed attempts to score by shooting from behind the free throw line. Each successful free throw is worth one point. \\ \hline
    Rebound & Obtaining the ball after a missed shot attempt. \\ \hline
    Foul & Attempting to unfairly disadvantage an opponent through certain types of physical contact. \\ \hline
    Turnover & A loss of possession by a player holding the ball. \\ \hline
    (Shot Clock) Violation & Failing to shoot the ball before the shot clock expires. Results in a turnover to the opponent team. \\ \hline
    Time-out & A limited number of clock stoppages requested by a coach or mandated by the referee for a short meeting with the players. \\ \hline
    Substitution & Replacing one player with another during a match. In basketball, substitutions are permitted only during stoppages of play, but are otherwise unlimited. \\\hline
    Jump Ball & A method used to begin or resume the game, where two opposing players attempt to gain control of the ball after an official tosses it into the air between them. \\ \hline
    Period &  NBA games are played in four periods (quarters) of 12 minutes. Overtime periods are five minutes in length. The time allowed is actual playing time; the clock is stopped while the play is not active. \\ \hline
    \end{tabular}
    \caption{Descriptions for Basketball terms used in our dataset. Explanations and rules derived from the official NBA rule-book at: \url{https://official.nba.com/rulebook/}, and the Basketball Wikipedia page at: \url{https://en.wikipedia.org/wiki/Basketball}}
    \label{tab:bball_dict}
\end{table}

For each event there are $10$ variables indicating the $5$ player lineup per team, which we use to monitor whether a player is on court at any given play. In a typical NBA game, there are about $450$ plays. Since we are only collecting data for key players, they are present on court during the vast majority of the game, totalling in an average of $337$ plays per player per game, for an average of $83$ plays per period.
For each period, we aggregate a player's performance through the following features: Points, Assists, Turnovers, Rebounds, Field goals made and missed, Free throws made and missed and mean and variance of shot distance from basket, for both successful and unsuccessful attempts. We build on these features to produce metrics that we believe capture the choice of actions made by the player (see Section~\ref{sec:task}).
Table~\ref{tab:metrics} provides each player's mean and standard deviation values for all performance metrics. The table also provides the average and standard deviation of the metrics across the entire dataset, information that we use to explain some of our findings in Section~\ref{sec:results} and modeling decisions in Section~\ref{sec:task}.

\subsection{Pre-game Interviews}
\label{subsec:interviews}
NBA players are interviewed by the press before and after games, as part of their contract with their team and with the league. The interviews take place on practice day, which is the day before the game, and on-court before, during and after the game. An NBA season has $82$ games per team, for all $30$ teams, spread across $6$ months, from October to April. Then, the top $8$ teams from each conference, Eastern and Western, advance to the playoffs, where teams face opponents in a knockout tournament comprised of a best-of-seven series. Playoff games gather much more interest, resulting in more interviews, which are more in-depth and with much more on the line for players and fans alike. Our dataset is hence comprised almost solely from playoff games data.

Interviews are open ended dialogues between an interviewer and a key player from one of the teams, with the length of the answers depending solely on the players, and the number of questions depending on both sides.\footnote{The most famous short response, by football player Marshawn Lynch, can be seen in: \url{https://www.youtube.com/watch?v=G1kvwXsZtU8}} Questions tend to follow on player responses, in an attempt to gather as much information about the player's state of mind as possible. For example: 

\begin{dialogue}

Q: "On Friday you spoke a lot about this new found appreciation you have this postseason for what you've been able to accomplish. For most people getting to that new mindset is the result of specific events or just thoughts. I'm wondering what prompted you specifically this off-season to get to this new mindset?"
\\

LEBRON JAMES: "It's not a new mindset. I think people are taking it a little further than where it should be. Something just -- it was a feeling I was after we won in Game 6 in Toronto, and that's how I was feeling at that moment. I'm back to my usual self."

\end{dialogue}

The degrees of freedom given, result in a significant variance in interview lengths. Sentences vary from as little as a single word to $147$, and interviews vary from $4$ sentences to $753$. Table~\ref{tab:interviewsperplayer} provides aggregated statistics about the interviews in our dataset, as well as the number of interviews, average number of Question-Answer (Q-A) pairs, average number of sentences and average number of words for each player.

In order to give further insight, we trained an LDA topic model~\cite{blei2003latent} for each player over all interviews he participated in, and present the top words of the most prominent topic per player in Table \ref{tab:topics}. Unsurprisingly, we can see that most topics involve words describing the world of basketball (f.e. game, play, team, championship, win, ball, shot) and the names of other players and teams, yet with careful observation we can spot some words relating to the player's or team's performance in a game (f.e. dynamic, sharp, regret, speed, tough, mental, attack, defense, zone). Generally, most topics contain similar words across players, yet some players show interesting deviations from the "standard" topic.\footnote{The LDA model is employed here for data exploration purposes only, specifically to show the general topic distribution per player in our dataset.}

\begin{table}[htbp]
    \centering
    \scriptsize
    \scalebox{0.96}{
    \begin{tabular}{|l|c|c|c|c|c|c|c|c|} \hline
    Player & PF & PTS & FGR & PR & SR & MSD2 & MSD3 & \# Plays \\ \hline
Al Horford & 2.07 & 12.5 & 0.52 & 0.26 & 0.22 & 5.97 & 6.81 & 299.14 \\ 
           & (1.07) & (6.71) & (0.18) & (0.19) & (0.17) & (4.74) & (6.69) & (57.97) \\ 
Andre Iguodala & 2.19 & 10.42 & 0.6 & 0.26 & 0.42 & 9.56 & 8.42 & 315.0 \\ 
               & (1.44) & (6.44) & (0.2) & (0.27) & (0.22) & (8.5) & (6.6) & (68.38) \\ 
Carmelo Anthony & 3.72 & 22.78 & 0.41 & 0.41 & 0.22 & 5.67 & 6.37 & 354.0 \\ 
                & (1.27) & (7.38) & (0.1) & (0.26) & (0.13) & (3.61) & (4.12) & (51.21) \\ 
Chauncey Billups & 2.84 & 18.42 & 0.5 & 0.23 & 0.41 & 13.7 & 9.54 & 368.61 \\ 
                 & (1.42) & (5.64) & (0.15) & (0.21) & (0.15) & (13.94) & (6.84) & (53.57) \\ 
Chris Bosh & 2.77 & 15.87 & 0.52 & 0.57 & 0.12 & 5.45 & 4.24 & 312.11 \\ 
           & (1.56) & (7.18) & (0.19) & (0.34) & (0.15) & (3.94) & (7.25) & (49.5) \\ 
Chris Paul & 3.43 & 19.8 & 0.51 & 0.22 & 0.31 & 9.81 & 9.72 & 341.14 \\ 
           & (1.17) & (7.43) & (0.1) & (0.15) & (0.14) & (5.27) & (6.45) & (57.69) \\ 
Damian Lillard & 2.08 & 27.42 & 0.48 & 0.43 & 0.39 & 10.51 & 9.19 & 370.9 \\ 
               & (1.24) & (8.07) & (0.1) & (0.25) & (0.13) & (5.73) & (5.23) & (39.93) \\ 
DeMar DeRozan & 2.73 & 26.0 & 0.53 & 0.48 & 0.08 & 5.71 & 2.12 & 345.45 \\ 
              & (1.19) & (10.52) & (0.09) & (0.21) & (0.09) & (1.58) & (3.64) & (30.37) \\ 
Derek Fisher & 3.13 & 8.67 & 0.54 & 0.26 & 0.36 & 13.06 & 7.46 & 282.97 \\ 
             & (1.61) & (5.04) & (0.22) & (0.21) & (0.22) & (14.27) & (8.26) & (60.32) \\ 
Dirk Nowitzki & 2.55 & 24.36 & 0.49 & 0.48 & 0.17 & 7.02 & 8.19 & 359.12 \\ 
              & (1.52) & (8.29) & (0.14) & (0.22) & (0.12) & (3.01) & (8.4) & (61.25) \\ 
Draymond Green & 4.0 & 13.02 & 0.54 & 0.3 & 0.39 & 8.49 & 7.34 & 359.68 \\ 
               & (1.38) & (6.59) & (0.19) & (0.18) & (0.16) & (9.61) & (6.75) & (47.89) \\ 
Dwyane Wade & 2.88 & 23.07 & 0.5 & 0.4 & 0.09 & 4.49 & 4.25 & 350.26 \\ 
            & (1.46) & (8.09) & (0.12) & (0.18) & (0.09) & (2.36) & (6.7) & (55.72) \\ 
James Harden & 2.9 & 23.8 & 0.47 & 0.32 & 0.43 & 9.5 & 8.13 & 351.27 \\ 
             & (1.58) & (9.69) & (0.14) & (0.19) & (0.14) & (6.08) & (6.01) & (66.4) \\ 
Kawhi Leonard & 2.56 & 13.0 & 0.47 & 0.46 & 0.33 & 8.51 & 8.03 & 277.17 \\ 
              & (1.69) & (5.3) & (0.18) & (0.32) & (0.15) & (6.82) & (6.08) & (70.58) \\ 
Kevin Durant & 2.58 & 28.34 & 0.54 & 0.44 & 0.29 & 8.92 & 9.75 & 378.85 \\ 
             & (1.44) & (6.86) & (0.11) & (0.24) & (0.09) & (4.13) & (5.69) & (52.26) \\ 
Kevin Garnett & 3.0 & 14.83 & 0.54 & 0.44 & 0.02 & 4.7 & 0.0 & 318.31 \\ 
              & (1.31) & (5.56) & (0.2) & (0.33) & (0.04) & (2.16) & (0.0) & (64.31) \\ 
Kevin Love & 2.33 & 15.75 & 0.44 & 0.47 & 0.44 & 11.0 & 8.13 & 281.58 \\ 
           & (1.34) & (9.17) & (0.15) & (0.27) & (0.16) & (7.95) & (4.79) & (58.15) \\ 
Klay Thompson & 2.51 & 19.34 & 0.48 & 0.45 & 0.48 & 16.17 & 8.92 & 352.96 \\ 
              & (1.49) & (8.65) & (0.11) & (0.28) & (0.15) & (14.97) & (4.18) & (58.37) \\ 
Kobe Bryant & 2.93 & 28.07 & 0.48 & 0.38 & 0.24 & 8.21 & 7.6 & 373.7 \\ 
            & (1.59) & (7.1) & (0.09) & (0.2) & (0.13) & (4.21) & (5.91) & (53.15) \\ 
Kyle Lowry & 3.33 & 21.56 & 0.49 & 0.3 & 0.48 & 14.09 & 8.94 & 334.67 \\ 
           & (1.32) & (9.9) & (0.14) & (0.2) & (0.17) & (11.57) & (3.97) & (46.15) \\ 
Kyrie Irving & 2.44 & 25.2 & 0.52 & 0.37 & 0.28 & 9.19 & 11.49 & 339.68 \\ 
             & (1.5) & (8.34) & (0.11) & (0.19) & (0.14) & (5.8) & (7.31) & (70.44) \\ 
Lamar Odom & 3.92 & 12.29 & 0.52 & 0.4 & 0.14 & 3.06 & 3.94 & 309.29 \\ 
           & (1.64) & (5.19) & (0.18) & (0.27) & (0.14) & (2.49) & (7.82) & (56.57) \\ 
LeBron James & 2.49 & 28.75 & 0.53 & 0.34 & 0.22 & 5.81 & 8.2 & 380.43 \\ 
             & (1.39) & (8.66) & (0.12) & (0.16) & (0.1) & (3.27) & (6.07) & (56.21) \\ 
Manu Ginobili & 3.02 & 14.84 & 0.5 & 0.39 & 0.44 & 9.6 & 7.98 & 276.0 \\ 
              & (1.24) & (7.47) & (0.19) & (0.24) & (0.14) & (7.8) & (5.76) & (65.68) \\ 
Pau Gasol & 2.97 & 16.54 & 0.56 & 0.36 & 0.01 & 2.72 & 0.0 & 356.95 \\ 
          & (1.2) & (5.99) & (0.14) & (0.25) & (0.03) & (1.88) & (0.0) & (70.03) \\ 
Paul George & 2.58 & 21.21 & 0.5 & 0.39 & 0.43 & 11.03 & 10.88 & 326.11 \\ 
            & (1.54) & (8.2) & (0.11) & (0.18) & (0.11) & (4.44) & (5.76) & (64.57) \\ 
Paul Pierce & 3.62 & 19.56 & 0.52 & 0.47 & 0.31 & 8.2 & 9.12 & 351.69 \\ 
            & (1.5) & (7.85) & (0.2) & (0.24) & (0.19) & (5.38) & (7.56) & (77.83) \\ 
Rajon Rondo & 2.74 & 12.19 & 0.5 & 0.23 & 0.08 & 3.22 & 3.97 & 354.26 \\ 
            & (1.51) & (6.53) & (0.21) & (0.11) & (0.09) & (2.27) & (8.03) & (59.83) \\ 
Ray Allen & 2.56 & 14.9 & 0.5 & 0.45 & 0.5 & 13.56 & 9.26 & 342.85 \\ 
          & (1.35) & (7.28) & (0.2) & (0.3) & (0.17) & (8.91) & (6.16) & (71.59) \\ 
Richard Hamilton & 3.61 & 20.83 & 0.5 & 0.36 & 0.07 & 5.1 & 2.32 & 382.09 \\ 
                 & (1.31) & (6.91) & (0.16) & (0.25) & (0.06) & (2.57) & (4.65) & (61.51) \\ 
Russell Westbrook & 2.85 & 24.3 & 0.45 & 0.33 & 0.21 & 5.16 & 6.46 & 363.52 \\ 
                  & (1.51) & (8.54) & (0.11) & (0.14) & (0.1) & (3.28) & (5.86) & (55.8) \\ 
Shaquille O'Neal & 3.68 & 15.79 & 0.61 & 0.71 & 0.0 & 1.43 & 0.0 & 275.37 \\ 
                 & (1.7) & (6.96) & (0.22) & (0.23) & (0.0) & (1.08) & (0.0) & (89.38) \\ 
Stephen Curry & 2.48 & 26.42 & 0.48 & 0.37 & 0.55 & 18.0 & 10.67 & 362.49 \\ 
              & (1.36) & (8.42) & (0.11) & (0.18) & (0.11) & (8.59) & (4.05) & (62.61) \\ 
Steve Nash & 1.73 & 19.09 & 0.55 & 0.25 & 0.24 & 8.42 & 8.93 & 334.45 \\ 
           & (1.2) & (6.38) & (0.13) & (0.15) & (0.12) & (3.77) & (7.0) & (46.15) \\ 
Tim Duncan & 2.65 & 18.43 & 0.51 & 0.46 & 0.01 & 2.55 & 0.44 & 324.07 \\ 
           & (1.31) & (7.53) & (0.15) & (0.28) & (0.03) & (1.41) & (3.27) & (58.05) \\ 
Tony Parker & 1.65 & 18.02 & 0.52 & 0.32 & 0.1 & 5.27 & 6.45 & 317.44 \\ 
            & (1.14) & (6.99) & (0.16) & (0.19) & (0.09) & (3.24) & (9.09) & (52.21) \\ 
  \hline
    dataset Average & 2.791 & 20.214 & 0.511 & 0.38 & 0.265 & 7.347 & 6.922 & 336.77 \\
    dataset Std. & 1.489 & 9.438 & 0.153 & 0.239 & 0.204 & 6.054 & 6.771 & 66.44 \\ \hline
    \end{tabular}}
    \caption{Performance metric mean and standard deviation (in parentheses) per player in our dataset. We report here the actual values of the performance metrics rather than deviations from the mean. For the definition of the performance metrics, see Section~\ref{sec:task}.}
    \label{tab:metrics}
\end{table}

\begin{table}[htbp]
    \centering
    \begin{tabular}{|l|c|c|c|c|} \hline
        Player & \# of & Avg. & Avg. & Avg.   \\
         & Interviews & \# of & \# of & \# of \\
         &  & Q-A pairs & sentences & words \\ \hline
        Al Horford & 14 & 6.43 & 45.29 &  685.4 \\ 
        Andre Iguodala & 26 & 12.69 & 111.08 & 1888.6 \\ 
        Carmelo Anthony & 18 & 14.28 & 86.78 & 1075.8 \\ 
        Chauncey Billups & 31 & 12.19 & 92.81 & 1545.6 \\ 
        Chris Bosh & 47 & 11.53 & 97.96 & 1311.2 \\ 
        Chris Paul & 35 & 12.49 & 82.66 & 1165.4\\ 
        Damian Lillard & 12 & 7.92 & 69.0 & 1115.5 \\ 
        DeMar DeRozan & 11 & 14.36 & 84.91 & 1259.1 \\ 
        Derek Fisher & 39 & 6.87 & 54.0 & 1149.7 \\ 
        Dirk Nowitzki & 33 & 12.15 & 114.39 & 1737.7\\ 
        Draymond Green & 59 & 14.97 & 140.88 & 2161.9 \\ 
        Dwyane Wade & 72 & 18.93 & 173.58 & 2494.7 \\ 
        James Harden & 30 & 10.73 & 63.77 & 858.6 \\ 
        Kawhi Leonard & 18 & 8.5 & 37.5 & 462.9 \\ 
        Kevin Durant & 67 & 17.73 & 138.25 & 2094.2 \\ 
        Kevin Garnett & 29 & 11.72 & 92.45 & 1459.1 \\ 
        Kevin Love & 24 & 11.79 & 89.54 & 1508.6 \\ 
        Klay Thompson & 53 & 13.42 & 112.25 & 1674.9 \\ 
        Kobe Bryant & 44 & 25.93 & 142.27 & 1861.1 \\ 
        Kyle Lowry & 9 & 14.44 & 92.67 & 1321.8 \\ 
        Kyrie Irving & 25 & 16.04 & 125.0 & 2377.6 \\ 
        Lamar Odom & 24 & 10.62 & 61.0 & 805.2 \\ 
        LeBron James & 122 & 22.6 & 189.43 & 2875.5 \\ 
        Manu Ginobili & 55 & 8.84 & 68.13 & 1032.8 \\ 
        Pau Gasol & 39 & 11.38 & 80.41 & 1281.6 \\ 
        Paul George & 19 & 12.58 & 88.63 & 1220.8 \\ 
        Paul Pierce & 32 & 15.28 & 122.78 & 1965.2 \\ 
        Rajon Rondo & 27 & 12.04 & 74.19 & 1021.8 \\ 
        Ray Allen & 39 & 8.31 & 63.36 & 1068.1 \\ 
        Richard Hamilton & 23 & 12.0 & 63.83 & 1099.2 \\ 
        Russell Westbrook & 40 & 18.48 & 108.05 & 1567.5 \\ 
        Shaquille O'Neal & 19 & 12.63 & 70.53 & 1043.8\\ 
        Stephen Curry & 71 & 17.7 & 156.92 & 2762.9 \\ 
        Steve Nash & 22 & 13.5 & 80.05 & 1132 \\ 
        Tim Duncan & 54 & 13.13 & 86.26 & 1380.9 \\ 
        Tony Parker & 55 & 12.75 & 80.84 & 1156.3 \\ \hline
        Dataset average & 37.14 & 14.52 & 110.28 & 16.38 \\ \hline
        Dataset standard deviation & 22.11 & 3.53 & 6.26 & 568.08 \\ \hline
    \end{tabular}
    \caption{Number of interviews and averages of number of Q-A pairs, sentences and words in an interview per player in our dataset.}
    \label{tab:interviewsperplayer}
\end{table}

\begin{landscape}
\setlength\tabcolsep{8 pt}
\begin{table}[htbp]
    \scalebox{0.8}{
    \begin{tabular}{|l|c|c|c|c|c|c|c|c|c|c|} \hline
Al Horford & live & angel & trip & basically & beautiful & allow & attack & next & week & league \\ \hline
Andre Iguodala & year & know & lot & rakuten & see & good & warrior & come & play & thing \\ \hline
Carmelo Anthony & game & tough & year & come & kobe & play & court & take & hand & back \\ \hline
Chauncey Billups & nba & teammate & award & twyman & thank & year & story & maurice & chauncey & applause \\ \hline
Chris Bosh & really & game & know & good & team & come & play & thing & want & look \\ \hline
Chris Paul & know & bowl & good & team & really & play & lot & shot & time & bowling \\ \hline
Damian Lillard & straight & breather & buckle & bad & begin & ne & steph & stage & sick & show \\ \hline
DeMar DeRozan & smith & suggest & tennis & talking & rival & skin & sit & sick & shut & shown \\ \hline
Derek Fisher & game & know & play & good & team & really & feed & come & thing & back \\ \hline
Dirk Nowitzki & good & great & team & time & back & game & lot & first & come & always \\ \hline
Draymond Green & game & thing & team & year & know & good & come & time & really & great \\ \hline
Dwyane Wade & game & team & play & know & good & year & come & time & last & feel \\ \hline
James Harden & game & good & know & shot & play & open & team & time & first & point \\ \hline
Kawhi Leonard & gear & matter & may & minute & morning & normal & noticing & opposite & order & padding \\ \hline
Kevin Durant & play & know & good & game & team & come & thing & shot & talk & want \\ \hline
Kevin Garnett & know & game & play & thing & lot & team & really & day & come & want \\ \hline
Kevin Love & game & team & year & play & lot & know & good & last & ball & feel \\ \hline
Klay Thompson & regret & scary & sharp & sharpness & shore & shrug & smith & speed & sulk & thigh \\ \hline
Kobe Bryant & game & good & play & night & come & take & really & something & much & talk \\ \hline
Kyle Lowry & challenge & curious & deep & dynamic & complete & contender & anything & cavalier & cake & bucket \\ \hline
Kyrie Irving & game & play & come & great & time & moment & tonight & team & big & would \\ \hline
Lamar Odom & really & win & year & know & last & happen & team & good & championship & would \\ \hline
LeBron James & know & game & year & team & last & able & time & play & take & thing \\ \hline
Manu Ginobili & game & know & tough & good & play & sometimes & see & thing & last & happen \\ \hline
Pau Gasol & zone & really & know & play & much & game & expect & tonight & obviously & sure \\ \hline
Paul George & know & team & something & feel & work & want & take & together & well & see \\ \hline
Paul Pierce & team & know & come & play & year & game & talk & look & really & lot \\ \hline
Rajon Rondo & game & great & ball & team & come & play & rebound & win & tonight & take \\ \hline
Ray Allen & standing & mental & marquis & orlando & operate & row & problem & thread & accustom & action \\ \hline
Richard Hamilton & relationship & resolve & demand & record & portland & phone & philly & pay & nut & new \\ \hline
Russell Westbrook & team & play & good & great & thing & able & game & come & time & different \\ \hline
Shaquille O'Neal & arena & city & fun & would & mistake & lot & talk & back & people & really \\ \hline
Stephen Curry & game & play & good & know & team & kind & year & really & time & obviously \\  \hline
Steve Nash & game & really & know & play & team & back & year & well & feel & win \\ \hline
Tim Duncan & game & play & good & team & time & come & back & lot & want & really \\  \hline
Tony Parker & good & game & play & never & rebound & chance & last & big & defense & keep \\  \hline
    \end{tabular}}
    \caption{Top 10 words in the most prominent topic for each player. A topic model was trained for each player on all his interviews in the dataset.}
    \label{tab:topics}
\end{table}
\end{landscape}
\section{The Task}
\label{sec:task}

Our goal in this section is to define metrics that reflect the player's in-game decisions and actions and formulate prediction tasks based on our definitions. Naturally, the performance of every player in any specific game is strongly affected by global properties such as his skills, and is strongly correlated with his performance in recent games. We hence define binary classification tasks, predicting whether the player is going to perform above or below his mean performance in the defined metrics. Across the dataset, we found that the difference between mean and median performance is insignificant and both statistics are highly correlated, hence we consider them as equivalent and focus on deviation from the mean.

Different players have significantly different variances in their performance (see Table~\ref{tab:metrics}) across different metrics. This phenomenon is somewhat inherent to basketball players due to the natural variance in player skills, style and position. Due to these evident variances, we did not attempt to predict the extent of the deviation from the mean, but preferred a binary prediction of the direction of the deviation. Another reason for our focus on binary prediction tasks, is that given our rather limited dataset size, and imbalance in number of interviews per player (some players were interviewed less than others, see Table~\ref{tab:interviewsperplayer}), we would like our models to be able to learn across players. That is, the training data for each player should contain information collected on all other players, pushing us toward a prediction task that could be calculated for players with a varying number of training examples and substantially different performance distributions.

\paragraph{Performance Metrics}

We consider 7 performance metrics: 
\begin{enumerate}
\setlength\itemsep{0.5em}
    \item \textit{FieldGoalsRatio(FGR)}
    \item \textit{MeanShotDistance2Points(MSD2)}
    \item \textit{MeanShotDistance3Points(MSD3)}
    \item \textit{PassRisk(PR)}
    \item \textit{ShotRisk(SR)}
    \item \textit{PersonalFouls(PF)}
    \item \textit{Points(PTS)}
\end{enumerate}
We denote with $M=\{FGR, MSD2, MSD3, PR, SR, PF, PTS\}$ the set of performance metrics. The performance metrics are calculated from the play-by-play data. In the notation below $p$ stands for a player, $t$ for a period identifier in a specific game and $\#$ is the count operator.\footnote{In the game dataset, $t$ denotes a specific game.} 
$\#\{event\}^{p,t}$ denotes the number of events of type $event$ for player $p$ in a game period $t$. We consider the following events: 

\begin{itemize}
    \item $shot$: A successful shot.
    \item $miss$: An unsuccessful shot.
    \item $2pt$: A two points shot.
    \item $3pt$: A three points shot.
    \item $assist$: A pass to a player that had a successful shot after receiving the ball and before passing it to any other player.
    \item $turnover$: an event in which the ball moved to the opponent team due to an action of the player.
    \item $pf$: a personal foul.
\end{itemize}

We further use the notations $Dist^{p,t}$ for a set which contains the distances for all the shots player $p$ took in period $t$, in which the distance of that shot from the basket is recorded, and $pts^{p,t}$ for the total number of points in a certain period. 
Our performance metrics, $m_{t}^{p}$, are defined for a player $p$ in a game period $t$, in the following way:

\begin{align}
MSD_t^{p} &= Mean(Dist^{p,t}) \\
\vspace{2cm}
PF_t^{p} &= \#\{pf\}^{p,t} \\
\vspace{2cm}
PTS_t^{p} &=  pts^{p,t} \\
\vspace{2cm}
FGR_t^{p} &= \frac{\#\{shot\}^{p,t}}{\# \{shot\}^{p,t}+\#\{miss\}^{p,t}} \\
\vspace{2cm}
SR_t^{p} &= \frac{\#\{3pt\}^{p,t}}{(\#\{miss\}^{p,t}+\#\{shot\}^{p,t})} \\
\vspace{2cm}
PR_t^{p} &= \frac{\#\{turnover\}^{p,t}}{ \#\{assist\}^{p,t}+\#\{turnover\}^{p,t}}
\vspace{2cm}
\end{align}
For MSD we consider two variants, MSD2 and MSD3, for the mean distance of 2 and 3 points shots, respectively.

\subsection{Prediction Tasks}

For each metric $m$ we define the player's mean as:
\begin{equation}
    \bar{m}^{p} =  \frac{\sum_{t=1}^{T^{p}}m^{p}_t}{|T^{p}|}
\end{equation}
where $T^{p}$ is the set of periods in which the player $p$ participated. We further define the per-metric label set $Y^m$ as:
\begin{equation}
  Y^m \! \!  = \! \{  y_t^{p,m} |
  p\in P, t\in T \}, y_t^{p,m} \! \!  = \! \! \left\{
  \begin{array}{@{}ll@{}}
    1,  &  m_{t}^{p} \geq  \bar{m}^{p}\\
    0,  & \text{otherwise}
  \end{array}\right.
\end{equation}
where $P$ is the set of players and $T$ is the set of periods.\footnote{Since $m_{t}^{p}$ is hardly ever equal to $\bar{m}^{p}$, the meaning of $y_{t}^{p,m} = 0$ is almost always a negative deviation from the mean.} 

For each player $p$ and period $t$, we denote with $x_{t}^{p}$ the player's interview text prior to the game of $t$, and with $y_{t}^{p,m}$ the label for performance metric $m$. In addition, lagged performance metrics are denoted with $y_{t-j}^{p,m}$, $\forall j\in\{1,2,...,k\}$ ($k=3$ in our experiments).\footnote{\textit{lagged performance metrics} refer to the same metric for the same player in the previous periods.}
We transform each sample in our dataset into interview-metric tuples, such that for a given player $p$ and period $t$ we predict $y_{t}^{p,m}$ given either:

\begin{enumerate}[label=(\alph*)]
\setlength\itemsep{0.5em}
    \item $x_{t}^{p}$: for the text-only mode of our models.
    \item $\{y_{t-j}^{p,m} | \forall j\in\{1,2,...,k\}, \forall m\in M\}$: for the metric only mode.
    \item $\{x_{t}^{p},y_{t-j}^{p,m} | \forall j\in\{1,2,...,k\}, \forall m\in M\}$: for the joint text and metric mode.
\end{enumerate}

While in this paper we consider an independent prediction task for each performance metric, these metrics are likely to be strongly dependent~\cite{vaz2012forecasting}. Also, we look at each player's actions independently, although there are connections between actions of different players and between different actions of the same player. We briefly discuss observed connections between our models for different tasks and their relation to the similarity between tasks in Section~\ref{sec:results}. However, as this is the first paper for our task, we do not attempt to model possible interactions between different players or between metrics which often occur in team sports, and leave these to be explored in future work.

\subsection{Performance Measures and Decision Making} Our paper is about the transmission of language into actions. In practice we try to predict performance metrics that are associated with such actions. Our measures aim to capture different aspects of the in-game actions made by players. FGR is a measure risk for the shots attempted. SR is also a measure of risk for attempted shots, yet it tries to capture a player's choice to take riskier shots that are worth more points. MSD2 and MSD3 are measures of the shot location, trying to capture for a given shot type (2/3 points) how far a player is willing to go in order to score. PR considers another offensive aspect, passes, and since it accounts for both turnovers and assists it captures part of the risk a player is willing to take in his choice of passing. PTS is a more obvious choice, it is the most commonly used metric to observe a player's offensive performance. PF is related to defensive decisions and is correlated with aggressive behavior.

By carefully observing the data presented in Table~\ref{tab:metrics} we can see that different metrics exhibit different levels of volatility across all players in our dataset.
More volatile metrics, such as field goals ratio (FGR), shot distance (MSD2/3) and shot risk (SR) are rather static at the player level but differ substantially between players. This volatility in shot related measures across players could be explained by the natural differences in shot selection between players in different positions. For example, back-court players generally tend to take more 3 point shots than front-court players. Events such as 3 point shots are therefore much sparser in nature for many players, and in many periods they occur at most once if at all. This causes the MSD3 (i.e. Mean Shot Distance for 3 point shots) to be $0$ many more times compared to other metrics in our dataset.
This volatility ultimately makes it harder to distinguish what drives variance in these metrics as opposed to more consistent metrics such as PF (Personal Fouls), PTS (Points) and PR (Pass Risk).

A possible explanation for PF and PTS being more consistent in our dataset is that they are considered rather critical performance measures to the overall teams' performance. Our dataset mainly consists of NBA All Stars (which are key players in their teams), interviewed before relatively important playoff games, and thus they are expected by their teams and fans to be more consistent in these critical measures.
While players differ substantially in terms of numbers of assists and turnovers, the pass risk (PR) metric accounts for this by looking at the ratio, resulting in a consistent measure across our dataset.

\section{Models}
\label{sec:models}

Our core learning task is to predict players in-game actions from their pre-game interview texts. Interviews are texts which contain a specific form of structured open-ended turn-based dialog between two speakers - the interviewer and the interviewee, which to a certain extent hold opposite goals in the conversation. Generally speaking, an interviewer's goal is to reveal pieces of exclusive information by giving the player a chance to reflect on his thoughts, actions and messages. However, the player's goal is to utilize the opportunity of public speaking to portray his competitive agenda and strengthen his brand, while maintaining a comfortable level of privacy. In-game performance metrics reflect different aspects of a player's in-game actions, which expose some information about the variance in a player's actions and performance between different games.

We formulated multiple binary classification tasks in Section~\ref{sec:task}, and these tasks pose several challenges from natural language processing perspectives:
\begin{itemize}
\setlength\itemsep{0.5em}
    \item \textit{Time Series}: Almost all samples in our data come from events (playoff series) which exhibit a certain form of time-dependence, meaning that subsequent events in the series may impact each other. This aspect requires careful treatment when designing our models and their features. 
    \item \textit{Remote Supervision Signal}: Our labels stem from variables (performance metrics) which are related to the speaker of the text and are only indirectly implied in the text. In this sense, our supervision signal refers to our input signal in an indirect and remote manner. This is in contrast to learning to predict the deviation from the mean based on past performance metrics, where the input and the output are tightly connected. This is also different from tasks such as sentiment analysis where the sentiment of the review is directly encoded in its text.
    \item \textit{Textual Structure}: Our input consists of interviews, which exhibit a unique textual structure of a dialog between two speakers, with somewhat opposing roles - an interviewer and an interviewee. We are interested in capturing information from these interviews, relevant to labels related only to the interviewee. Yet, it is not trivial to say whether this information appears in the interviewee's answers alone or what type of context and information do the interviewer's questions provide. 
\end{itemize}
In light of these challenges, we design our models with $4$ main questions in mind: 
\begin{enumerate}
\setlength\itemsep{0.5em}
    \item Could classification models utilize pre-game interview text to predict some of the variance in players' in-game performance at both game and period levels?
    \item Could text be combined with past performance metrics to produce better predictions?
    \item How could we explicitly model the unique textual structure of interviews in order to facilitate accurate performance prediction?
    \item Could Deep Neural Networks (DNNs) jointly learn a textual representation of their input interview together with a task classifier to help us capture textual signals relevant to future game performance?
\end{enumerate}

To tackle these questions, we chose to design metric-based, text-based and combined models, and assign the $-M$, $-T$ and $-TM$ suffixes to denote them respectively. Within each set of models, we chose to explore different modeling strategies in an increasing order of complexity and specialization to our task. We next provide a high level discussion of our models, and then proceed with more specific details.

\paragraph{Metric-based models}
We implement two standard autoregressive models, which are commonly used tools in time-series analysis, alongside a BiLSTM~\cite{hochreiter1997long} model. Both models make a prediction for the next time step (game/period) given performance metrics from the three previous time steps. These models exhibit the predictive power of performance metrics alone, and serve as baselines for comparison to text-based models.

\paragraph{Text-based models}
We design our text-based models to account for different levels of textual structure. We start by implementing a standard Bag-of-Words text classifier which represents an interview as counts of unigrams. We continue by implementing a word-level CNN~\cite{lecun1995convolutional} model, which represents interviews as a sequence of words in their order of appearance. We then implement a sentence-level BiLSTM model, which represents interviews as a sequence of sentences, where each sentence is represented by the average of its word embeddings. Finally, we chose to implement a BERT~\cite{devlin2019bert} model, which accounts for the interview structure by representing interviews as sequences of question-answer pairs. Each pair's embeddings are learned jointly by utilizing the model's representations for pairs of sequences, which are based on an attention mechanism~\cite{vaswani2017attention} defined over the word-level contextual embeddings of the question and the answer. This serves as an attempt to account for the subtler context a question induces over an answer, and for the role of each speaker in the dialog. These text-based models exhibit the predictive power of text alone in our prediction task.

\paragraph{Combined models}
DNNs tranform their input signals into vectors and their computations are hence based on matrix calculations. This shared representation of various input signals makes theses models highly suitable to multi-task and cross-modal learning, as has been shown in a variety of recent NLP works (e.g.~\cite{Sogaard:16, Rotman:18, Malca2018neural}). We therefore implemented variants of our best performing LSTM and BERT text-based models which incorporate textual features from the pre-game interview with performance metrics from the previous three time steps. These models help us quantify the marginal effect of adding textual features in predicting the direction of the deviation from the player's mean performance, over metric-based models.
We next describe each of our models in details.

\subsection{Metric-based Autoregressive Models}

An autoregressive (AR($k$)) model is a representation of a type of a random process. It is a commonly used tool to describe time-varying processes, such as player performance. The AR model assumes that the output variable ($y_{t}^{p,m}$) depends linearly on its own $k$ previous values and on a stochastic term $\epsilon_{t}$ (the prediction error)~\cite{akaike1969fitting}. We focus on AR($3$) to prevent loss of data for players with very few examples (previous games) in our dataset:
\begin{equation}
\label{eq:ar-m}
    y_{t}^{p,m} = c + \sum_{j=1}^3 \varphi_{j} y_{t-j}^{p,m} + \epsilon_{t}
\end{equation}
We also consider using all lagged metrics as features for predicting a current metric:
\begin{equation}
\label{eq:mstar}
    y_{t}^{p,m} = c + \sum_{j=1}^3 \sum_{w \in M} \varphi_{j}^{w} y_{t-j}^{p,w} + \epsilon_{t}
\end{equation}
That is, we make predictions for a given game $t$, player $p$ and metric $m$, based on performance in the previous $k=3$ games, using either the same metric $m$ (Equation~\ref{eq:ar-m}) or all metrics in $M$ (Equation~\ref{eq:mstar}).

We employed a standard linear regression and a logistic regression. We tested both models for all $k$ values for which we had enough data, and $k=3$ was chosen since it performed best in development data experiments. We report results only for the linear regression model, since both models performed similarly.

\subsection{The BoW and TFIDF Text Classifiers}

The bag-of-words (BoW) and term frequency-inverse document frequency (TFIDF)~\cite{salton1991developments} models are standard for text classification tasks~\cite{Yogatama:14}, and they therefore serve as our most basic text-based models. We constructed both BoW and TFIDF feature vectors per interview, and have tried using unigrams and bigrams, alone or in combination. We considered Random Forest (RF)~\cite{liaw2002classification}, Support Vector Machine (SVM)~\cite{cortes1995support} and Logistic Regression (LR)~\cite{ng2002discriminative} classifiers. While BoW provides a straight-forward effective way to represent text, it assumes n-gram (in our case we tried $n=1, n=2, n=1,2$) independence and therefore does not take the structure of the text into account. TFIDF adjusts for the fact that some words are more frequent in general, but makes the same assumptions. We report results for the Random Forest (RF) classifier for both BoW (unigrams) and TFIDF (unigrams + bigrams) feature sets, since these consistently performed better in development data experiments, for BoW and TFIDF respectively. Finally, since these simple models were consistently outperformed by our best text-based DNN models (see Section ~\ref{sec:results}), we did not attempt to incorporate any performance metrics as features into them.

\subsection{Deep Neural Networks}

DNNs have proven effective for many text classification tasks \cite{Kim:14,Ziser:18}. An appealing property of these models is that training a DNN using a supervision signal results not only in a predictive model, but also with a representation of the data in the context of the supervision signal. This is especially intriguing in our case, where the supervision signal\com{(decisions and actions taken in the following game)} is not clearly visible in the text, and is more related to its speaker.

Moreover, the text in our task is structured as a dialog between two speakers, which entails an additional level of contextual dependence between speakers, on top of the internal linguistic structures of the utterances produced by the individual speakers. These factors pose a difficult challenge from a modeling perspective, yet DNNs are known for their architectural flexibility which allows learning a joint representation for more than one sequence~\cite{chen2016thorough}, and have shown promising performance in different tasks where models attempt to capture nuanced phenomena in text~\cite{peters2018elmo}.

We consider three models that excel on text classification tasks: CNN~\cite{Kim:14}, BiLSTM~\cite{hochreiter1997long} and BERT~\cite{devlin2019bert}. In order to obtain a vectorized representation of an interview's text, we employed different text embedding techniques per model, each based on different pre-trained embedding models. 
Below we describe the various models.

\subsubsection{The CNN Model}

\paragraph{Motivation}
We implement a standard word-level CNN model for text classification (CNN-T), closely following the implementation described in~\cite{Kim:14}. This model showed promising results on various text classification tasks such as sentiment classification and stance detection~\cite{Kim:14}.
By implementing this model we aim to examine the extent to which a standard word-level text classification neural network, which does not explicitly account for any special textual structure except for the order of the words in the text, can capture our performance metrics from text.

\paragraph{Model Description} 
Interviews are fed into the model as a sequence of words in their order of appearance in the interview. We concatenated the interview's word embedding vectors into an input matrix, such that embeddings of consequent words appear in consequent matrix columns. Since interviews vary in length, we padded all word matrices to the size of the longest interview in our dataset.
We then employed three 2D convolution layers with max-pooling and a final linear classification layer.

\subsubsection{The BiLSTM Models}

\paragraph{Motivation}
Our CNN model treats an interview as a single sequence of words, and apart from the fact that the order of words is maintained in the input matrix, it does not model any textual structure. By implementing BiLSTM-based models we aim to directly model the interview as a sequence of sentences, rather than of words.
We believe that since interviews involve multiple speakers interacting in the form of questions and answers, where each question and answer are comprised of multiple sentences, a sequential sentence-level model could capture signals word-level models cannot.

\begin{figure}[htbp]
    \centering
    \includegraphics[scale=0.11]{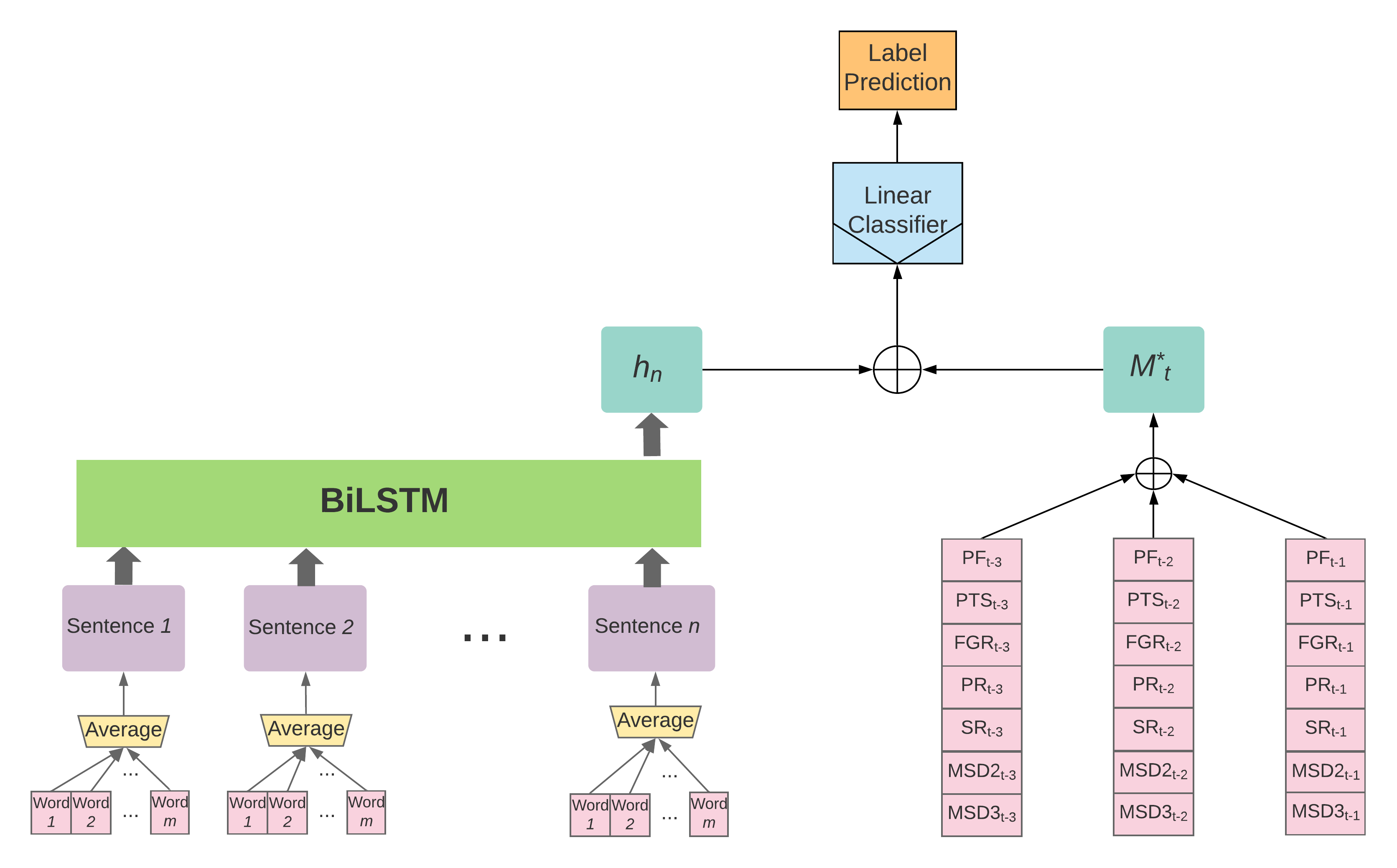}
    \caption{The LSTM-TM Model. $h_n = h_{n}^{forward} \oplus h_{n}^{backward}$, $|h_n| = |h_{n}^{forward}|+|h_{n}^{backward}|$. $\oplus$ denotes the vector concatenation operator.}
    \label{fig:lstm-tm}
\end{figure}

We chose to implement our text-based BiLSTM (LSTM-T) as a sentence-level sequential model, where each sentence is represented by the average of its pre-trained word embeddings~\cite{adi2016fine}. Since BiLSTM is a general sequential model, it also fits naturally as an alternative time-series model for performance metrics only (LSTM-M), similar to the AR($k$) model described in Equation~\ref{eq:mstar}. The various model variants allow us to examine the independent effects of text and metrics on our prediction tasks, using the same underlying model. Moreover, we can now examine the effect of combining text and metric features together in a BiLSTM model (LSTM-TM) by concatenating the metric feature vectors used as input to LSTM-M with the final textual vector representation produced by LSTM-T (see Figure~\ref{fig:lstm-tm}).

\paragraph{Model Description}
We next provide the technical implementation details of each of our BiLSTM-based models.


\paragraph{LSTM-T} The BiLSTM model for text is fed with the sentences of the interview, in their sequential order. Each sentence is represented by the average of its word embeddings. The BiLSTM's last hidden-state forward and backward vectors are concatenated and fed into two linear layers with dropout and batch normalization, and a final linear classification layer (see the left part of Figure~\ref{fig:lstm-tm}).

\paragraph{LSTM-M} The BiLSTM model for metrics, which mimics the AR($k=3$) model of Equation~\ref{eq:mstar}, is fed at each time step with the performance metric labels from the last three time steps. We concatenate the last hidden states (forward and backward) and feed the resulting vector as input to a linear classifier. This model is almost identical to LSTM-T, differing only in the input layer.

\paragraph{LSTM-TM} The BiLSTM model which combines text and metrics utilizes a similar mechanism as LSTM-T to produce the text vector representation. We then concatenate a vector containing all metrics from the past three time steps, to the text vector. The resulting vector is fed into a binary classifier, similar to the one described for LSTM-T (see Figure~\ref{fig:lstm-tm}).

\subsubsection{The BERT Models}

\paragraph{Motivation} 
We are seeking to capture information regarding the player's pre-game state through the interview text, which is comprised of a series of consecutive Question-Answer pairs. In an interview, a player controls only his answers, where his choice of language can be observed in the context of the questions he is asked. While the player does not have any control over the questions, these can be viewed as a second order approximation of the player's state since an interviewer is purposefully phrasing the questions directed at the player. Alternatively, one can view the questions as external information which we cannot attribute to the player based solely on that fact. We choose to proceed with the former approach, to view the questions as a valuable context to the player's answers.

Since the unique structure of an interview encourages a form of speaker roles and contextual dependence, which may seem similar to other "looser" forms of discourse, in this work we choose to focus our modeling on the local dependencies within each pair of a question and its immediate answer. In future work we plan to further explore the interview structure in our modeling.

\paragraph{Interview Representation}
Our CNN model treats an interview as a single sequence of words, while our BiLSTM model treats an interview as a single sequence of sentences where each sentence is represented by the average of its word embeddings. Both these models do not take into account any other characteristics of the interview structure. While there are CNN and LSTM-based models which aim to capture document structure, for example hierarchy~\cite{yang2016hierarchical}, adapting these to capture the subtleties of an interview structure is a non-trivial task.


BERT provides a method for producing a single joint contextual representation for two related text sequences (such as Question-Answer (Q-A) pairs), which attempts to represent both texts and the relations between them. We found this feature useful for our task and a natural fit for modeling interview structure, as it allowed us to break up each interview to its Q-A pairs, input them in sequence to BERT, and produce a respective sequence of Q-A vectors. We follow a similar method for producing Q-A vectors as described in the BERT paper~\cite{devlin2019bert} and provide further details in our model description below.

From a technical perspective, handling texts which greatly vary in length\com{and have a high average length of~1800 words}, requires some creativity, if we do not want to lose data by truncating long texts to a fixed maximum length. Moreover, it is empirically shown that many recurrent models, for example LSTM, suffer performance degradation with increasing sequence length~\cite{luong2015effective}. Our choice of breaking up each interview into its Q-A pairs allows us also to handle shorter sequences at the interview level, as opposed to longer sentence or word level sequences\com{(each interview is comprised of~14.5 Q-A pairs on average, as opposed to~1800 words or~110 sentences on average)}. We let BERT carry out the heavy task of handling the word sequences for each Q-A pair,\com{(average of~125 words per Q-A pair, 99th percentile: 356 words)} since it can handle a sequence of up to 512 tokens. We hypothesize that these factors contribute to a more effective interview representation.

\begin{figure}[htbp]
    \centering
    \includegraphics[scale=0.11]{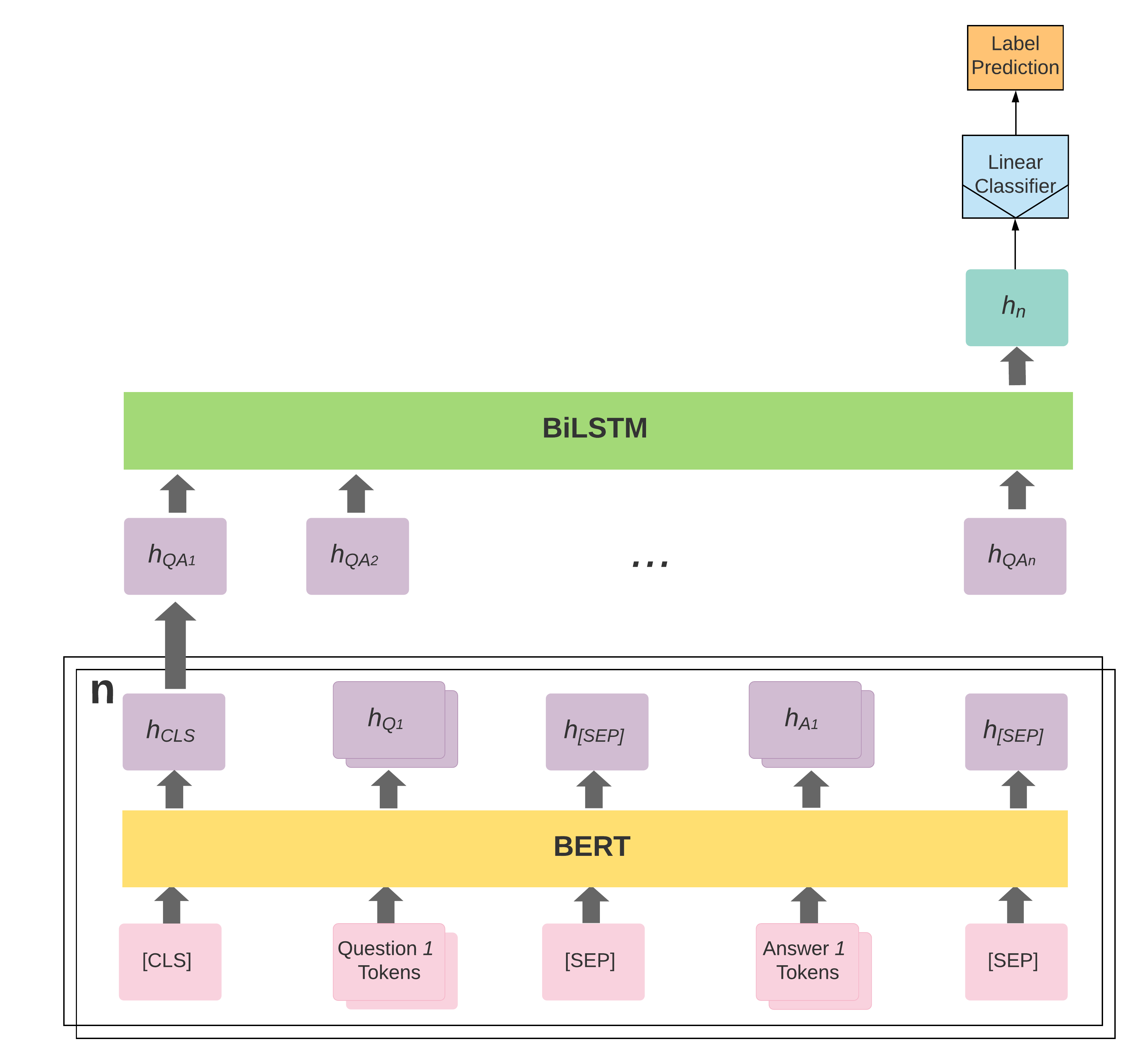}
    \caption{The BERT-L-T Model. $n$ denotes the number of Q-A pairs in a given interview. Each Q-A pair is fed into BERT to produce a Q-A vector, and the resulting vectors are then fed in sequence to the BiLSTM. $h_n$ is generated in the same way as in the LSTM-TM model (see Figure~\ref{fig:lstm-tm}).}
    \label{fig:bert-l-t}
\end{figure}

\paragraph{Model Description} 
Interviews are fed into the BERT model as a sequence of Question-Answer (Q-A) pairs in their order of appearance in the interview. We follow the terminology and methodology presented in the BERT paper~\cite{devlin2019bert}, which considers the $[CLS]$ token vector from the last hidden layer of the BERT encoder as a representation of an entire input sequence for classification tasks. When a single input is comprised of two sequences of text (a Q-A pair in our case), $text\_A$ represents the first sequence (a Question in our case) and $text\_B$ represents the second sequence (an Answer in our case). Each sequence ends with the special $[SEP]$ token, which represents the end of a single sequence and acts as a separator between the two sequences. For each Q-A pair we produce a Q-A vector, by extracting the vector associated with the special $[CLS]$ token from the last hidden layer of the BERT encoder. This results in a sequence of Q-A vectors per interview (see Figure~\ref{fig:bert-l-t}).

To produce a single vector representation per interview we implement two alternative models, BiLSTM (BERT-L-T) and Attention (BERT-A-T), both are detailed below. The final interview vector is fed into a linear classification layer activated with a Sigmoid function~\cite{finney1952sigmoid}, to produce a binary prediction. At training, the BiLSTM parameters (BERT-L-T) and the attention parameters (BERT-A-T) are jointly trained with the classifier parameters. In both cases we employ a pre-trained BERT model as a source of text representation (BERT feature-based approach~\cite{devlin2019bert}\footnote{The pre-trained model was downloaded from: \url{https://github.com/google-research/bert}.}), and do not fine tune its text representation or classification parameters on our data, to avoid heavy computations.

\paragraph{BERT-L-T} 
A BiLSTM is sequentially fed with the Q-A vectors, and its last hidden states (forward and backward) are concatenated to serve as the interview vector. See Figure~\ref{fig:bert-l-t} for an illustration of the model architecture.

\begin{figure}[htbp]
    \centering
    \includegraphics[scale=0.11]{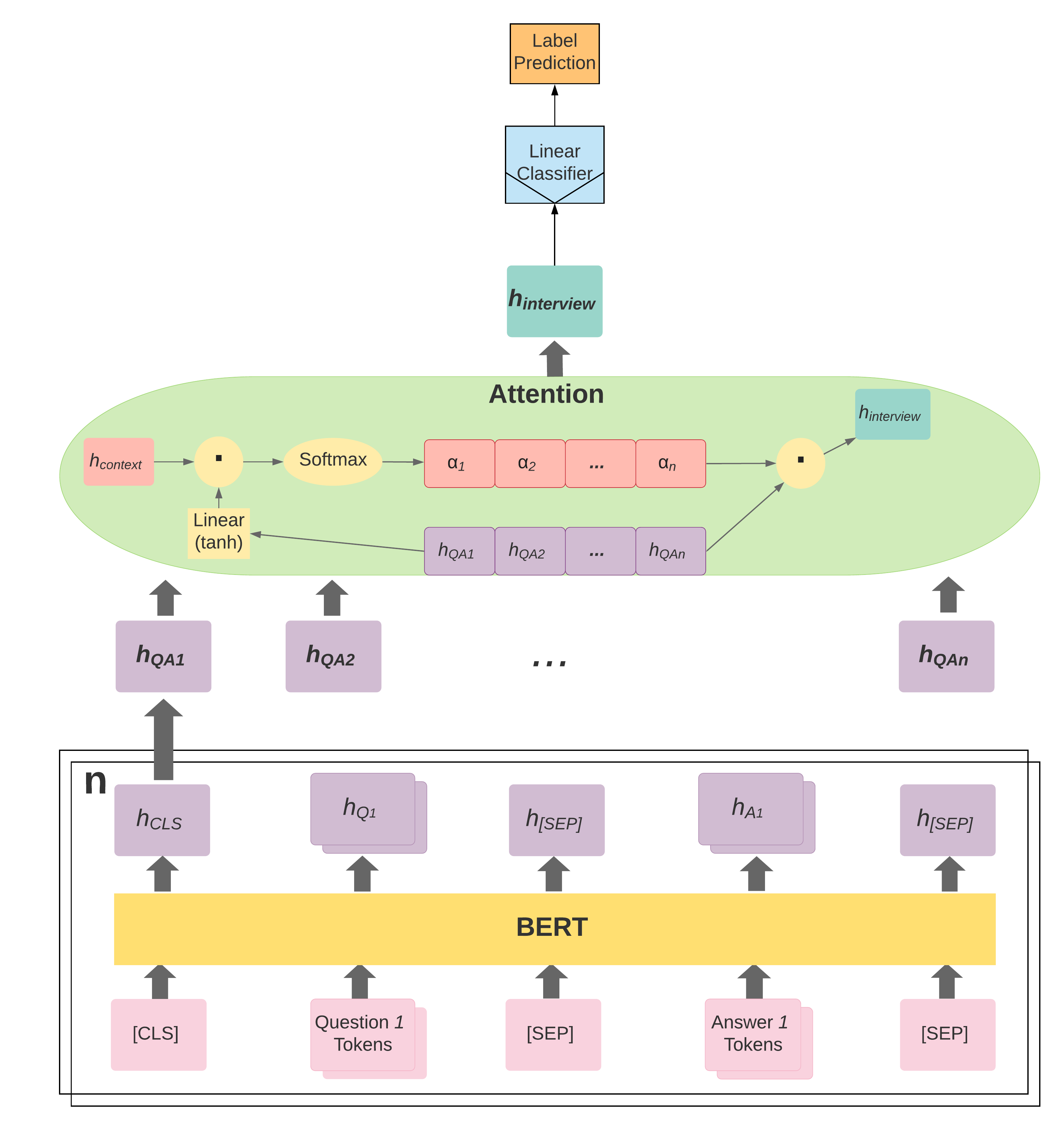}
    \caption{The BERT-A-T Model. Attention is applied over a sequence of Q-A vectors, which are produced by feeding the interview's Q-A pairs into BERT. $h_{context}$ is randomly initialized and jointly learned with the attention weights during the training process.}
    \label{fig:bert-a-t}
\end{figure}

\paragraph{BERT-A-T} 
A simple Attention mechanism (described in~\cite{yang2016hierarchical}) is employed over the sequence of Q-A vectors, and produces a pooled vector which serves as the interview representation. See Figure~\ref{fig:bert-a-t} for an illustration of the model architecture. \com{This model is very similar to the BERT-L-T model shown in Figure~\ref{fig:bert-l-t}, except that the top BiLSTM layer is replaced by an attention layer.} In almost all of our experiments BERT-A-T and BERT-L-T performed similarly, yet the BERT-A-T model proved to be slightly but consistently superior (see Section \ref{sec:results}). We hypothesize that the attention mechanism serves as an efficient and effective method of pooling our Q-A vectors, which results in a much lighter model in terms of the number of learned parameters.

\begin{figure}[htbp]
    \centering
    \includegraphics[scale=0.11]{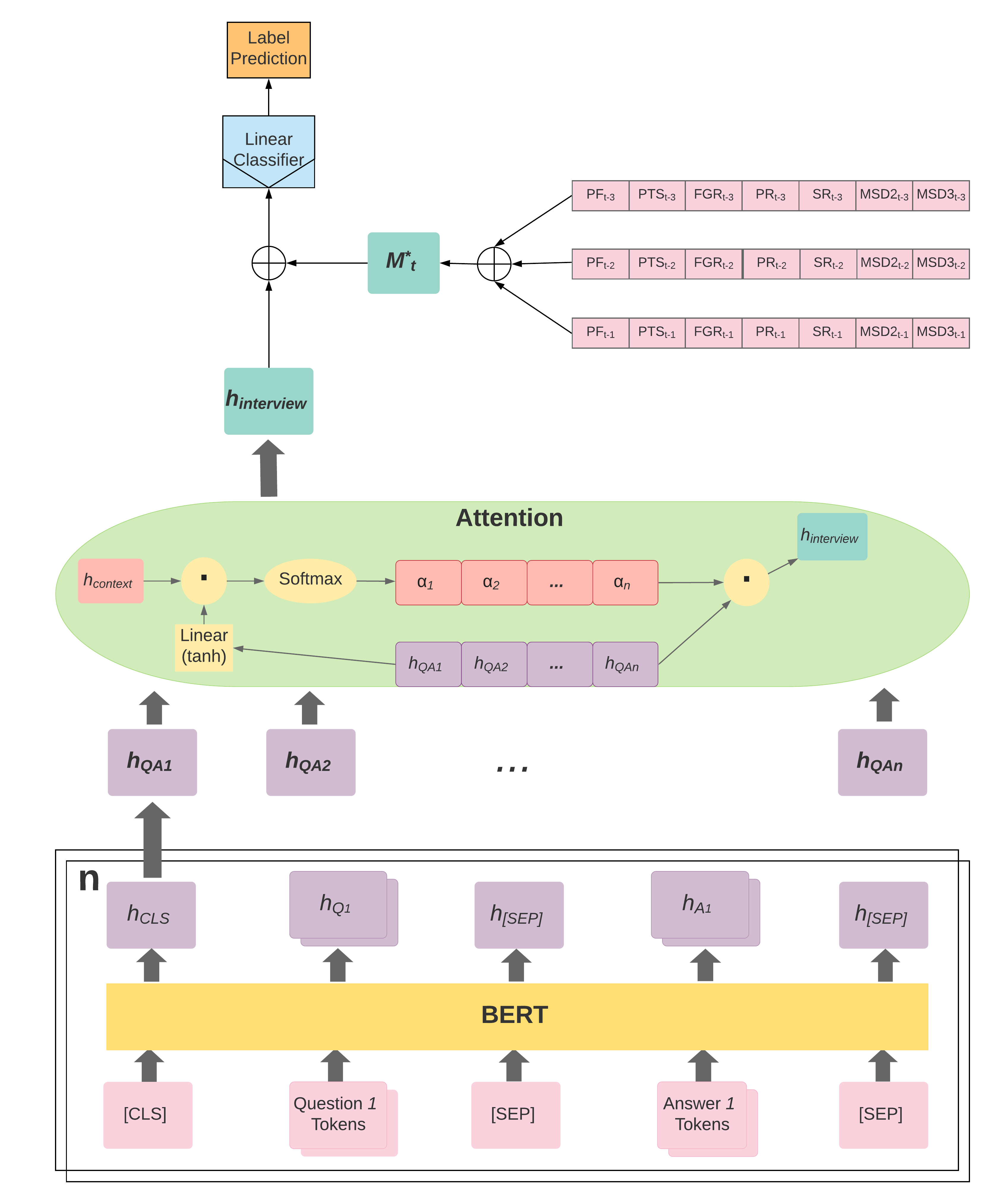}
    \caption{The BERT-A-TM Model. We use the same notation as presented in Figure~\ref{fig:bert-a-t}.}
    \label{fig:bert-a-tm}
\end{figure}

\paragraph{BERT-A-TM}
We implemented a variant of the BERT-A-T model, where before the interview vector is fed into the classifier, it is concatenated with all performance metric labels from the last three games. This model lets us explore the combined value of textual and performance metric signals. See Figure~\ref{fig:bert-a-tm} for an illustration of the model architecture.

\section{Experiments}
\label{sec:experiments}

\subsection{Tasks and Data}
We perform two sets of experiments, differing in the level of metric aggregation: (a) game-level; and (b) period-level.\footnote{Recall that each game is comprised of 4 periods.} Our period-level task does not distinguish between different periods within a game (that is, the model does not distinguish between, e.g., the second and the third period). In order to solve this task we hence train a classifier on data aggregated at the period-level from the various periods in our dataset. We experiment with both levels of aggregation in order to explore different aspects of the players' actions and how they manifest in different parts of the game. While game-level data is less volatile and can capture more general differences in a player's performance, it could fail to show behavioral fluctuations that are more subtle, such as "clutch" decisions, momentum performance boosts or a short series of mistakes. The period-level data can catch those subtleties and tell a more fine-grained story, though it is more sensitive to rare events, such as 3 point shots and fouls. 

The differences between these two sub-tasks are demonstrated by the per-metric label distributions (see Table~\ref{tab:metrics}). Events such as 3 point shots are sparser in nature -- many periods they occur at most once if at all. This causes the MSD (i.e. Mean Shot Distance) of those shots to be $0$ and leads to extreme class imbalance, making the classification task a lot more difficult.

Events such as shots in general, as captured by the PTS (Points) and FGR (Field Goal Ratio) metrics, occur more regularly, and thus result in balanced classes at both the game and period levels. Balanced classes are generally desired in binary classification tasks, since imbalanced classes could easily bias models towards the common class in the training data, making it almost impossible for us to determine whether the models captured even the slightest effects from the data. This is especially desired in light of research question ($\#4$) presented in Section~\ref{sec:models}, where we set a goal to understand whether DNNs could learn a textual representation capable of capturing textual signals for our tasks. We would hence like to avoid the potential effects of imbalanced classes, which could inhibit our models from learning such textual representations. 

The question of balanced data also stems in our task with respect to the interviewed players, since we do not have an equal number of interviews for all players (see Table~\ref{tab:interviewsperplayer}). This could potentially bias our models towards specific players which are more prevalent in the dataset. This could also complicate splitting our dataset into training, development and test sets. For each subset, we would ideally prefer to maintain the same ratio of interviews per player as in the entire dataset, in addition to maintaining the same positive to negative classes ratio.

In this study, we chose to employ a stratified 5-fold cross validation process (see Section~\ref{subsec:cv} below), in order to maintain the positive to negative class ratio across our training, development and test subsets. We did not attempt to explicitly maintain the ratio of interviews per player, since our aforementioned stratified process yielded subsets which fairly maintained this ratio. In future research, we plan on exploring the effects of different interviews per player ratios, to examine whether certain players exhibit linguistic or performance patterns different than other players, and whether our models could capture such patterns or be biased by them.

\subsection{Models}
We consider the following models (described in further detail in Section~\ref{sec:models}), and use the $-T$, $-M$ and $-TM$ suffixes to denote 
model variants for textual, metric and combined features, respectively:
\begin{itemize}
    \item \textit{AR($3$)-M} - a linear autoregressive model, which considers the last three time steps of the predicted performance metric.
    \item \textit{AR($3$)-M*} - a linear autoregressive model, which considers the last three time steps of all performance metrics.
    \item \textit{LSTM-M} - a BiLSTM model which considers the last three time steps of all performance metrics.
    \item \textit{BoW-RF-T} - a Random Forest classifier which utilizes a unigram bag-of-words features set.
    \item \textit{TFIDF-RF-T} - a Random Forest classifier which utilizes a TFIDF feature set defined over unigrams and bigrams.
    \item \textit{CNN-T} - a word-level CNN model.
    \item \textit{LSTM-T} - a sentence-level BiLSTM model.
    \item \textit{LSTM-TM} - a model similar to LSTM-T, except that the text representation is combined with the last three time steps of all performance metrics, and the result is fed to the classification layer.
    \item \textit{BERT-L-T} - a model that explicitly accounts for the Q-A structure of the input interviews, with BERT representations and LSTM sequence modeling.
    \item \textit{BERT-A-T} - a model that explicitly accounts for the Q-A structure of the input interviews, with BERT representations and an attention mechanism.
    \item \textit{BERT-A-TM} - a model similar to BERT-A-T, except that the text representation is combined with the last three time steps of all performance metrics, and the result is fed to the classification layer.
\end{itemize}

Recall our four research questions from Section~\ref{sec:models}. 
Our experiments are designed to compare between text and metric-based models, demonstrating the predictive power of text-based models in our task. In addition, they are designed to highlight the effects of different modeling strategies, in an increasing order of complexity and specialization to our task. 
Finally, we compare to the common class (CC) baseline which assigns to every test set example the most common training label. We chose to add this baseline in order to examine the performance of our models in comparison to a more naive and "data-driven" approach which does not model either text nor past metrics~\cite{sim2016friends}. 

\subsection{Cross Validation}
\label{subsec:cv}
We randomly sampled $20\%$ of our interviews and generated a held-out test set for each performance metric, per game and period tasks, each consisting of interviews and their related performance metrics.\footnote{Recall that in our period-level task we do not distinguish between the different periods within a game.}
We then implemented a 5-fold cross validation procedure for each metric label, in each fold randomly sampling $80\%$ of the remaining interviews for training and $20\%$ for development. All our training, development and test sets are stratified: the ratio of positive and negative examples in each subset is identical to the ratio in the entire dataset.\footnote{We achieved this by utilizing the StratifiedShuffleSplit and StratifiedKFoldCV utility methods from \textit{scikit-learn}, using a random seed of $212$.} 

\subsection{Implementation Details and Hyperparameters}

All models were developed in Python, utilizing different packages per model.
\paragraph{Autoregressive Models}
We developed all models utilizing the \textit{statsmodels} package~\cite{seabold2010statsmodels}.
\paragraph{Bag-of-Words Models}
We developed all models with \textit{scikit-learn}~\cite{Pedregosa2011scikit}.

\subsubsection{Deep Neural Network Models}
\label{subsec:DNN_hyper}
For all Neural Network models, we used Dropout~\cite{srivastava2014dropout} with $p=0.2$ and batch normalization for linear layers, ReLU as the activation function for all internal layers, and Sigmoid as the activation function for output layers.
Training is carried out for $500$ epochs with early stopping and a batch size of $8$ samples (interviews). Due to the variance in sentence and interview length, we employed various batch padding (to the maximum length in batch) and masking techniques.
We used binary cross entropy as our loss function, and the ADAM optimization algorithm~\cite{DBLP:journals/corr/KingmaB14} with the parameters detailed in Table~\ref{tab:adam_params}.

\begin{table}[htbp]
    \centering
    \begin{tabular}{|c|c|} \hline
       Parameter & Value \\ \hline
       Learning Rate & $5e^{-04}$ \\ 
       Fuzz Factor $\epsilon$ & $1e^{-08}$ \\ 
       Learning rate $decay$ over each update & $0.0$ \\  \hline
    \end{tabular}
    \caption{The ADAM optimizer hyper-parameters.}
    \label{tab:adam_params}
\end{table}

\paragraph{The CNN-T Model}
We employ GloVe word embeddings~\cite{pennington2014glove}, trained on the 2014 Wikipedia dump + the Gigaword 5 corpus ($6B$ tokens, $400K$ word types, uncased) where each word vector is of dimension $d=100$.\footnote{\url{http://nlp.stanford.edu/data/glove.6B.zip}}
We developed this model with Keras~\cite{chollet2015keras} over TensorFlow~\cite{abadi2016tensorflow}. The hyper-parameter values of the model are given in Table~\ref{tab:cnn_params}.
\setlength{\tabcolsep}{0.5em}
\FloatBarrier
\begin{table}[htbp]
    \centering
    \begin{tabular}{|c|c|} \hline
       Layer & Filter Size \\ \hline
       Convolution 1 & $3$ \\ 
       Convolution 2 & $4$ \\ 
       Convolution 3 & $5$ \\  
       Linear Output & $1$ \\ \hline
    \end{tabular}
    \caption{The CNN-T model hyper-parameters.}
    \label{tab:cnn_params}
\end{table}
\FloatBarrier

\paragraph{The BiLSTM Models}
For our text-based BiLSTM models (LSTM-T and LSTM-TM), we employ the same GloVe word embeddings as in the CNN model described above. The size of the hidden textual representations at the forward and backward LSTMs is $100$.
Our LSTM-M model's hidden state vector size is $7$ since we have $|M|=7$ metrics. 
We developed these models with PyTorch~\cite{paszke2017pytorch}. The hyper-parameter values of the model are given in Table~\ref{tab:lstm_params}.
\begin{table}[htbp]
    \centering
    \begin{tabular}{|c|c|c|} \hline
       Layer & Input Size & Output Size \\ \hline
       Input (Embedding) & $|Vocabulary|$ & $100$ \\
       $LSTM^{forward} \oplus LSTM^{backward}$ & $100$ & $200$ \\
       Linear 1 & $200$ & $100$ \\ 
       Linear 2 & $100$ & $32$ \\ 
       Linear Output & $32$ & $1$ \\  \hline
    \end{tabular}
    \caption{The LSTM-T model hyper-parameters}
    \label{tab:lstm_params}
\end{table}
\paragraph{BERT Models}
For our BERT models, we utilize BERT's pre-trained models as a source of text representation.
We experimented with two uncased pre-trained BERT models, both trained on the BookCorpus ($800M$ words)~\cite{zhu2015bookcorpus} and Wikipedia ($2,500M$ words): \textit{BERT-Base} ($L=12$ layers, $H=768$ hidden vector size, $A=12$ attention heads, $P=110M$ parameters) and \textit{BERT-Large} ($L=24$, $H=1024$, $A=16$, $P=340M$), both publicly available via source code provided by Google Research's GitHub repository.\footnote{\url{https://github.com/google-research/bert}}
The \textit{BERT-Large} model slightly outperformed \textit{BERT-Base} in all of our experiments, hence we report results only for \textit{BERT-Large}.
We developed these models with PyTorch~\cite{paszke2017pytorch}, utilizing and modifying source code from HuggingFace's "PyTorch Pretrained BERT" GitHub repository.\footnote{\url{https://github.com/huggingface/pytorch-pretrained-BERT}}
Table~\ref{tab:bert-l_params} details the hyper-paramters used for the BERT-L-T model and Table~\ref{tab:bert-a_params} details the hyper-parameters used for the BERT-A-T and BERT-A-TM models.
\begin{table}[htbp]
    \centering
    \begin{tabular}{|c|c|c|} \hline
       Layer & Input Dimensions & Output Dimensions \\ \hline
       BERT Pretrained Encoder & Interview text & $H$ x \# Q-A pairs\\
       $LSTM^{forward} \oplus LSTM^{backward}$ & $H$ x Max \# Q-A pairs in batch & $2H$ \\
       Linear Output & $2H$ & $1$ \\  \hline
    \end{tabular}
    \captionsetup{justification=centering}
    \caption{The BERT-L-T model hyper-parameters. \\ 
    $H$ is the pre-trained BERT model's hidden vector size ($H_{base}=768$, $H_{large}=1024$)}
    \label{tab:bert-l_params}
\end{table}

\begin{table}[htbp]
    \centering
    \begin{tabular}{|c|c|c|} \hline
       Layer & Input Dimensions & Output Dimensions \\ \hline
       BERT Pretrained Encoder & Interview text & $H$ x \# Q-A pairs \\
       Attention & $H$ x Max \# Q-A pairs in batch & $H$ \\
       Linear Output & $H$ & $1$ \\  \hline
    \end{tabular}
    \captionsetup{justification=centering}
    \caption{The BERT-A-T model hyper-parameters. \\
    $H$ is defined as in Table~\ref{tab:bert-l_params}.}
    \label{tab:bert-a_params}
\end{table}

We further experiment with our BERT models, by continuing the Language Model pre-training process for both the \textit{BERT-base} and \textit{BERT-large} uncased pre-trained models, on the interviews from our dataset.
Our goal in this experiment is to evaluate whether further pre-training of BERT on interview data would yield text representations which better capture features relevant to the basketball domain and hopefully improve prediction performance on our tasks.

We utilized the standard \textit{Masked Language Model} (MLM) and \textit{Next Sentence Prediction} (NSP) pre-training objectives of BERT (see ~\cite{devlin2019bert}). 
We ran the pre-training process for $1$ and $3$ epochs on the interview texts at the sentence level (to accommodate the NSP task), and tuned all BERT layers in this process. 
After completing the pre-training process, we used the new pre-trained BERT models as part of new varaints of BERT-A-T and BERT-A-TM, and evaluate their performance on all $7$ tasks for both the game and the period levels. We denote these models as \textit{BERT-EPT-A-T} and \textit{BERT-EPT-A-TM} respectively, where EPT stands for "extended pre-training". 
Our results indicate that the BERT models with extended pre-training are less effective than the standard BERT models that are not pre-trained on interview text. We hence report our results with the standard BERT models and analyze the extended pre-training process in Section \ref{sec:results}.
\section{Results}
\label{sec:results}
Examining and analyzing our results, we wish to address the four research questions posed in Section~\ref{sec:models}. That is, we wish to assess the interviews' predictive power without and alongside past metrics (questions $\#1$ and $\#2$, respectively), the benefit of modeling the interviews' textual structure (question $\#3$) and the ability of DNNs to learn a textual representation relevant for predicting future performance metrics (question $\#4$).

\paragraph{Overview}
The results are presented in Table \ref{tab:results} (top: game-level, bottom: period-level). 
First, they suggest that pre-game interviews have predictive power with respect to performance metrics on both game and period level tasks (question $\#1$). This is evident by observing that text-based ($-T$) models generally performed better than the most common class baseline (CC) and metric-based ($-M$) models. Performance for all BERT-based and LSTM-based models is superior to CC and metric-based models at the game-level, yet at the period-level results are rather mixed.
Second, they suggest that combining pre-game interviews with past performance metrics yields better performing models (question $\#2$). This can be seen in the performance gain of our combined ($-TM$) models over their respective text-based models, and the overall best performance of the BERT-A-TM model in most tasks.
Third, they support the use of structure-aware DNNs for these prediction tasks (question $\#3$). This can be seen by the general performance gain of text-based models as their modeling complexity of textual structure rises, especially in game-level tasks. Furthermore, our DNN models generally outperformed non-neural models, suggesting that DNNs are able to learn a textual representation suitable to our tasks (question $\#4$).
We shall examine the results in further detail below, in light of our four research questions.

\setlength{\tabcolsep}{0.4em}
\begin{table}[htbp]
    \scalebox{1.3}{
    \begin{tabular}{|l|c|c|c|c|c|c|c|} \hline
        Model & PF & PTS & FGR & PR & SR & MSD2 & MSD3 \\ \hline
        CC  & 50.4 & 52.4 & 50.9 & 52.8 & 57.3 & 55.3 & 58.9 \\ \hdashline
        AR(3)-M  & 43.7 & 49.6 & 48.9 & 51.9 & 57.3 & 53.7 & 59 \\
        AR(3)-M*  & 48.1 & 54.9 & 51.5 & 47.8 & 57.5 & 55.3 & 56.3 \\
        LSTM-M & 50.1 & 50.5 & 48.2 & 48.8 & 58.2 & 56.9 & 55.9 \\ \hdashline
        BoW-RF-T & 50.8 & 50.8 & 51.1 & 53.7 & 57.8 & 54.9 & 59.3 \\
        TFIDF-RF-T & 51.5 & 53.7 & 53.3 & 53.3 & 57.6 & 58.0 & 57.6 \\
        CNN-T  & 51.9 & 54.9 & 50.4 & 48.5 & 57.5 & 53.7 & 57.8 \\
        LSTM-T & 56.7 & 58.2 & 55.7 & 58.2 & 55.2 & 55.8 & 58.2 \\
        BERT-L-T & 54.9 & 55.6 & 55.6 & 57.5 & 59.5 & 57.8 & 60.5 \\
        BERT-A-T & 57.5 & 60.1 & 58.7 & 58.6 & \textbf{63.4} & \textbf{61.9} & 60.8 \\ \hdashline
        LSTM-TM & 57.5 & 58.2 &  55.5 & 57.7 & 61.6 & 59.7 & 55.8 \\
        BERT-A-TM & \textbf{59.3} & \textbf{60.7} & \textbf{60.5} & \textbf{60.1} & 61.2 & 59.7 & \textbf{60.8} \\ \hline
    \end{tabular}}
\newline
\vspace*{3mm}
\newline
    \scalebox{1.3}{
    \begin{tabular}{|l|c|c|c|c|c|c|c|} \hline
        Model & PF & PTS & FGR & PR & SR & MSD2 & MSD3 \\ \hline
        CC & 52.2 & 54.1 & 51.5 & 64.8 & 64.3 & 64 & 75.1 \\ \hdashline
        AR(3)-M  & 52.6 & 53.6 & 48.8 & 65 & \textbf {64.9} & 65.7 & 75.9  \\
        AR(3)-M*  & 53.3 & 54.2 & 51.7 & \textbf{65} & 63.2 & \textbf{65.7} & \textbf{75.9} \\
        LSTM-M & 50.1 & 51.5 & 50.3 & 62.7 & 62.4 & 62.3 & 74.2 \\ \hdashline
        BoW-RF-T & 52.3 & 52.5 & 49.3 & 60 & 61.6 & 58.2 & 67.1 \\
        TFIDF-RF-T & 50.2 & 53.9 & 50.3 & 63.1 & 63.5 & 63.7 & 74.5 \\
        CNN-T  & 51.6 & 53.3 & 49.4 & 59.2 & 62.5 & 57 & 68.1 \\
        LSTM-T & 51.6 & 53.7 & 52.3 & 64.1 & 64.9 & 63.1 & 74 \\
        BERT-L-T & 53.2 & 56 & 53.6 & 63.8 & 63.6 & 64.3 & 74.5 \\
        BERT-A-T & 55.6 & 53.7 & 52.3 & 64.3 & 63.7 & 64.2 & 75.1 \\ \hdashline
        LSTM-TM & 52.3 & 58.3 & 56.6 & 63.6 & 63.4 & 63.4 & 74.2 \\
        BERT-A-TM & \textbf{56.3} & \textbf{58.5} & \textbf{57.8} & 64.1 & 64.7 & 64.9 & 74.7 \\
         \hline
    \end{tabular}}
    \caption{Game-level (top) and period-level (bottom) accuracy on a 0-100 scale. Best accuracy on each performance metric is highlighted in bold. CC stands for the most common class baseline. Models that involve text perform best in all game-level tasks and in 3 of 7 period-level tasks.} 
    \label{tab:results}
\end{table}

\paragraph{The Predictive Power of Interviews}
Game-level BERT-A-T, our top performing text-based model, outperforms the CC baseline and all metric-based models, in all $7$ tasks, with improvements over the CC baseline ranging up to an added accuracy of: $7.1\%$ on personal fouls (PF), $7.7\%$ on points (PTS), $7.8\%$ on field goal ratio (FGR), $5.8\%$ on pass risk (PR), $6.1\%$ on shot risk (SR), $6.6\%$ on mean 2-point shot distance (MSD2) and $1.9\%$ on mean 3-point shot distance (MSD3).
Game-level LSTM-T outperforms the most common class baseline (CC) and all metric-based models in $5$ of the $7$ tasks.

For four period-level tasks, PR, SR, MSD2 and MSD3, the common class baseline and metric-based models outperform text-based and combined models. We hypothesize that this is because at many periods the participation of many of the players is limited, which results in a below average performance. For all tasks at the game-level, as well as for PF, PTS and FGR at the period-level, this is much less frequent and the text is then much more informative.

Interestingly, for most game-level tasks metric-based models (that rely only on performance metric information), are not able to predict much better than a coin flip or the most common class baseline. Only in three cases a metric-based model outperforms these baselines by more than $1\%$: game-level PTS, period-level PF and period-level MSD2. These results emphasize the achievement of text-based prediction that succeeds where standard approaches fail.

\paragraph{The Predictive Power of Text and Performance Metric Combination}
BERT-A-TM outperforms BERT-A-T, its text-only counterpart which outperforms all other models across all $7$ game-level tasks, in $4$ out of these $7$ tasks (PF, PTS, FGR and PR, they are on par for MSD3) and in $5$ out of $7$ period-level tasks.
Game-level BERT-A-TM performs particularly well on the personal fouls (PF) and pass risk (PR) tasks, outperforming the most common class baseline (CC) by $8.9\%$ and $7.3\%$, respectively. Interestingly, both metrics tend to be similar across players (see Table~\ref{tab:metrics}). Deviations from the mean in these metrics reflect an increased or decreased level of aggressiveness (PF for defensive and PR for offensive decisions), suggesting that this quality is somewhat more visible in the language shown in the interviews.

More volatile metrics, such as field goals ratio (FGR), shot distance (MSD2/3) and shot risk (SR) are rather static at the player level but differ substantially between players (see Table~\ref{tab:metrics}).
This makes it harder to distinguish what drives variance in those metrics. Our results on those tasks are not as strong as those of the PF, PTS and PR tasks, but besides MSD3 we still observe significant improvement for most text-based ($-T$) and combined models ($-TM$) over the alternative models (see game-level results in Table~\ref{tab:results}).

A closer look into the data reveals a potential explanation for the superiority of BERT-A-TM on period-level PF, PTS and FGR: Since these performance metrics are less volatile between periods, it is useful for BERT-A-T to employ past performance metrics to better predict current player performance, compared to the text-only alternative.\footnote{All three performance metrics have an average standard deviation of $0.7$ at the period-level, substantially below the other four performance metrics which exhibit an average standard deviation of $1.4$.} Adding performance metric information is hence particularly useful in these setups.

The game-level LSTM-TM model substantially outperforms the LSTM-T model, its text-only counterpart, on SR and MSD2, but the two models perform similarly on the other tasks, with the exception of MDS3, where LSTM-T outperforms LSTM-TM. Period-level LSTM-TM outperforms LSTM-T on PF, PTS, and FGR, yet LSTM-T outperforms LSTM-TM on SR. Overall, combining text and metrics with the LSTM-based models proves to be valuable in some of the tasks, but certainly not in all of them.
Given the comparison between BERT-A-TM and BERT-A-T, and between LSTM-TM and LSTM-T, we can conclude that BERT better captures the combination between text and past performance metrics compared to LSTM.

We also look at the variation in our models' performance across all 5 folds, to test for their robustness. For this, we calculated the standard deviation of our models across all folds and have found that in all $7$ game level tasks and $7$ period level tasks, the average standard deviations of BERT-A-T and BERT-A-TM were lower than the average standard deviations of each of the non-neural models (BoW-RF-T, TFIDF-RF-T and the AR models). On average, the standard deviation of the of the BERT-A-T and BERT-A-TM models is $1\%$ (absolute, i.e. $1$ accuracy point), a fifth of the standard deviation of the TFIDF-RF-T, the best-performing non-neural model. Some non-neural models are less noisy than TFIDF-RF-T, but have a standard deviation of at least $1.5\%$ and are still much noisier than the than the BERT-A-T and BERT-A-TM models.

To summarize, BERT-A-TM is our overall best performing model, outperforming all other models in a total of $8$ (five game-level and three period-level tasks) out of $14$ tasks. In all seven game-level prediction tasks and in three of the period-level tasks (in total $10$ out of $14$ tasks), it is a BERT-A model that performs best. In eight of these cases it is BERT-A-TM and only in two it is BERT-A-T. This clearly indicates the added value of the textual signal on top of the signal in past metrics. 

\paragraph{The Value of DNN Modeling: Textual Structure and Representation Learning}
With regards to our third and fourth questions, Table~\ref{tab:results} reveals a general performance gain when using text-based DNN models as they capture more intricate aspects of the interview structure. Notably, at both the game and the period levels the text-based BERT-A-T model outperforms the TFIDF-RF-T non-neural text-based model (in $7$ of $7$ game level comparisons and in $6$ of $7$ period level comparisons). Moreover, it is clear from the table that overall the text-based BERT-A-T model outperforms the text-based LSTM-T model, which in turn outperforms the text-based CNN-T model. This leads to two conclusions. First, it suggests that DNNs are able to learn a textual representation suitable for our tasks, which might be needed due to the remote nature of the supervision signal, as discussed in Section~\ref{sec:models} (question $\#4$). Second, it suggests that modeling the textual structure of interviews is valuable for our tasks (question $\#3$).

Particularly, modeling an interview on a question-answer level better captures the interview structure compared to simpler word or sentence-level modeling. In addition, we hypothesize that the attention mechanism handles sequences of jointly represented question-answer pairs better than more rigid sequential models such as LSTMs. This observation is further supported by the better performance of the attention-based BERT-A-T model compared to BERT-L-T, which employs an LSTM instead of attention on top of the BERT representations (BERT-A-T performs better in $7$ of $7$ game level tasks, and both models perform similarly on the period-level tasks). The above observations clearly demonstrate the benefits of modeling an interview's textual structure, in an increasing level of complexity and nuance with respect to the unique characteristics of interviews.
%

\paragraph{Extended BERT Pre-training} When examining the BERT-EPT-A-T and BERT-EPT-A-TM models, we observe a slight yet consistent degradation in prediction performance throughout almost all tasks, in comparison to both the BERT-A-T and the BERT-A-TM models. There is an average accuracy degradation of $0.6\%$ over all tasks: $0.7\%$ on the game level, and on the period level the results are mostly on par, with the exception of PTS, FGR and SR metrics where BERT-EPT-A-TM is somewhat better. Moreover, we observe an increase in overall result variance, when comparing our BERT-EPT-A models (average standard deviation of $1.8\%$) to the BERT-A models (average standard deviation of $1\%$). This suggests that the extended pre-training process yields less stable models.
We hypothesize that continuing the pre-training process might lead to BERT overfitting on our data, at the cost of "forgetting" valuable features from its original pre-trained representations. This could be explained by the following observations: 
\begin{enumerate}
    \item Our dataset size is orders of magnitude smaller (about $2M$ words in total) compared to the BookCorpus ($800M$ words) and Wikipedia ($2,500M$ words) datasets on which BERT was originally pre-trained.
    \item Our pre-trained models which ran for $1$ epoch outperformed our pre-trained models which ran for $3$ epochs, by an average of $0.2\%$. This suggests that longer pre-training would slightly increase the "forgetting" effect and harm prediction performance.
    \item The NSP task might be inadequate for interview text, since it does not allow for capturing the unique structures present in interviews, such as the dependence between speakers' utterances, which are usually comprised of multiple sentences each.
\end{enumerate}

In light of our experimental results, and specifically our observation on the possible limitations of the NSP task for interview processing, we believe that investigating the process of pre-training BERT for interview text is an interesting direction for future work.
For example, one idea is to replace the standard NSP task with a similar task, designed for Q-A pairs, with the goal of predicting whether a given Q and A are a real pair in an interview.
In future work, we would like to further explore different possible pre-training methods which might be better-suited for interviews, by designing them to capture the unique structures present in interviews and the roles of the participating speakers.

\begin{figure}[htbp]
\centering
\includegraphics[width=\textwidth]{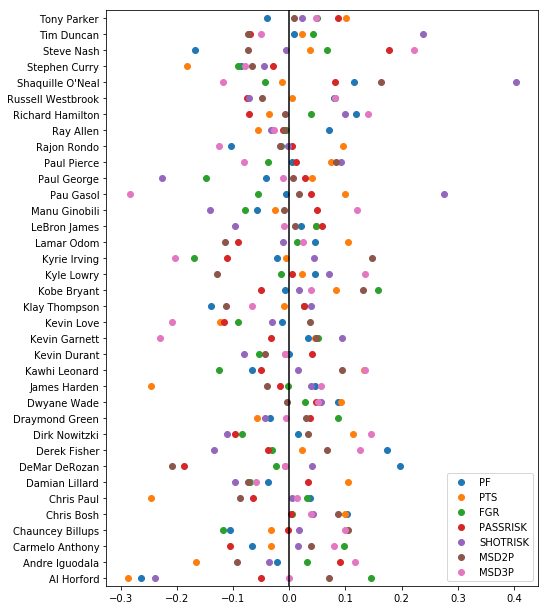}
\caption{BERT-A-T prediction accuracy per player, relative to its accuracy for all players, for each prediction task. Each point in the graph is defined by Equation~\ref{eq:diff_acc}.}
\label{fig:pred_acc}
\end{figure}

\paragraph{Per-player Model Performance}
In the analysis so far we have adhered to results on aggregated performance, without observing differences between players. To provide an analysis of the model's performance per player, we present in Figure \ref{fig:pred_acc} the relative performance of the BERT-A-T model at the game-level, for each player and for each task. We shall use the definitions below to explain the figure structure:
Let BERT-A-T's accuracy score for a given metric $m \in M$, over all examples in the test set $I_{test}$, be defined as:
\begin{equation}
    ACC^{m} = \frac{\sum_{i \in I_{test}} \mathbbm{1} \{ \hat{y}_{i}^{m} = {y}_{i}^{m} \}}{|I_{test}|}
\end{equation}
where $\hat{y}_{i}^{m}$ denotes the label predicted by BERT-A-T and ${y}_{i}^{m}$ denotes the true label for example $i$.
Let BERT-A-T's accuracy score for a given metric $m \in M$ and a player $p \in P$, over the subset of all test set examples $I^p_{test}$ in which player $p$ participates, be defined as:
\begin{equation}
    ACC^{p,m} = \frac{\sum_{i \in I^p_{test}} \mathbbm{1} \{ \hat{y}_{i}^{m} = {y}_{i}^{m} \}}{|I^p_{test}|}
\end{equation}
Each point in Figure~\ref{fig:pred_acc} is defined as the difference between $ACC^{p,m}$ and $ACC^m$:
\begin{equation}
\label{eq:diff_acc}
    \Delta ACC^{p,m} = ACC^{p,m} - ACC^m
\end{equation}
We use this difference measure to facilitate the visualization of BERT-A-T's relative performance in terms of accuracy for all players in all tasks, on a single graph.

Observing this figure, some interesting patterns arise. First, model performance in SR prediction is significantly better for front-court players, e.g. Tim Duncan, Shaquille O'Neal and Pau Gasol, that make significantly fewer attempts to shoot behind the 3-point arc, in a manner that could be quite predictable.\footnote{Court structure, including the location of the 3-point arc, is presented in Figure~\ref{fig:shot_loc}.}  Second, the performance of BERT-A-T is significantly lower in predicting the distance of the 3 point shots taken by those players (the average of this distance is captured by the MSD3 metric). This could be the result of the infrequent nature of these types of shots by these players. Comparatively, performance of back-court players, such as Kyle Lowry, Manu Ginobili and Steve Nash, who frequently attempt 3 point shots, are better predicted by our BERT-A-T model.
\section{Qualitative Analysis}
\label{sec:qual}

Thus far we have shown the predictive power of our models, but we would also like to get an insight as to what in the text is indicative of the player's actions and decisions. Relying on the sports psychology and analytics literature, we would like to test the efficacy of theories such as IZOF~\cite{hanin1997emotions} and its successors (see discussion of such theories in Sections~\ref{sec:intro} and~\ref{sec:related}). That is, we would like to observe whether the success of our models in predicting players' actions is related to the players' hypothesized "emotional state" as reflected in the text.

The rise of deep learning models, such as those we employ here, has produced substantially better prediction models for a plethora of tasks, most notably for those that rely on unstructured data~\cite{peters2018elmo, devlin2019bert}. This improved predictive power, however, has come at the expense of model explainability and understandability. Understanding predictions made by deep neural networks is notoriously difficult, as their layered structure which is coupled with non-linear activations does not allow for tracing and reasoning over the effect of each input feature on the model's output. In the case of text-based models this problem is amplified, as features are usually comprised of adjacent sequences of words, and not abstract linguistic concepts that might exist in the text and push the model towards specific predictions.

Following the meteoric rise of such models, there have been many attempts to build tools that allow for model explanation and interpretation~\cite{NIPS2017_7062,ribeiro2016should}. However, such tools often rely on local perturbations or shallow correlations. They are ill-suited for text-based models and for reasoning on higher level concepts, such as our players' states. When attempting to reason about abstract concepts such as "emotional state", it is not clear how to highlight the effects of different features on the predictions a model makes, when its input features are word sequences. To be able to estimate such effects we need to represent the models' output in a space where we can reason about high level concepts. Specifically, we decided to represent such concepts as interview topics.

\subsection{LDA Topic Modeling for DNN Interpretation}
Following the above reasoning, we qualitatively interpret the predictions of our models using Latent Dirichlet Allocation (LDA)~\cite{blei2003latent}. Importantly, we do not perform predictions with LDA, but instead use the topics it induces in order to interpret the results of our DNN models. To the best of our knowledge, the idea of interpreting the predictions of DNNs in text classification tasks using LDA topics has not been proposed previously.
LDA models the interviews as mixtures of topics, where each topic is a probability distribution over the vocabulary. Our goal in this analysis is to find the topics that are most associated with the predictions our DNN models make in each of the seven performance classification tasks. Our reasoning is that LDA-based topics may provide a latent space suited for intuitive reasoning about the predictions of DNN models, allowing enough dimensions for observing differences in higher level concepts, while keeping it relatively compact for proper analysis.

Interpreting a supervised DNN prediction model utilizing an auxilliary unsupervised LDA model, which itself learns a latent representation of the data, raises a question about the predictive power of the LDA model in the original prediction task. Suppose the LDA topics prove to hold the same predictive power as the DNN or they are perfectly correlated with the DNN's predictions, then the LDA model could serve as an interpretable alternative to the uninterpretable DNN. This is not the usual case with most DNN prediction models, which generally outperform LDA-based counterparts in most text-based prediction tasks.
Therefore, we are interested in finding LDA topics that correlate with the DNN predictions, keeping in mind that we need to quantify this correlation in order to understand how well the topics explain the DNN predictions.
We chose to perform two types of analyses:
\begin{itemize}
    \item Associating a specific topic with the positive predictions of the analyzed classifier, for each prediction task (see Table~\ref{tab:top-topics} and Figure~\ref{fig:topics_preds}). This yields a rather informative and intuitive analysis, which sheds light on what a prediction model managed to consistently capture in the text. Yet, the selection of a single topic might limit the interpretation to a single specific aspect and miss other aspects.
    \item Analyzing the correlations between all LDA topics and the predictions of the analyzed classifier, for each performance task (see Figure~\ref{fig:heat_map}). This analysis is complementary to the previous one: it is multi-dimensional and hence gives a higher level view rather than delving into the details of one explaining topic.
\end{itemize}
We describe each analysis in further detail below.

\subsection{Associating Topics with Classifier Predictions}

We train a topic model\footnote{Using the \textit{gensim} library~\cite{rehurek2010gensim}.} and optimize the number of topics on our entire training set $I_{train}$ for maximal coherence~\cite{lau2014machine}, resulting in a set $Z$ of $36$ topics. Then, for each label we split the test set $I_{test}$ into positively and negatively classified interviews, according to the BERT-A-T predictions, our best performing text-based model. We note that as we are interested in interpreting our model's predictions, we perform the split according to the model predictions and not according to the gold standard.

Following this split, we then search for the topic that is on average most associated with the model's positive predictions, i.e. the topic with the largest probability difference between positive and negative model predictions. We denote the topic distribution (mixture) of an interview $i$ with $\theta^i$ (as defined in~\cite{blei2003latent}), and $p(z|\theta^i) = \theta_z^i$ as the probabilities of words in interview $i$ to be generated by topic $z$. Intuitively, $\theta_z^i$ can be thought of as the degree to which topic $z$ is referred to in interview $i$~\cite{blei2001lda}.
The computations for both positive and negative topics are described in Equations~\ref{eq:pos_topic} and~\ref{eq:neg_topic}.

\begin{equation}
\label{eq:pos_topic}
    PositiveClassTopic^m = z^{m+} = \argmax_{z \in Z} \{ \sum_{i \in I_{test}} \hat{y}_{i}^{m}\cdot \theta_z^i \}
\end{equation}

\begin{equation}
\label{eq:neg_topic}
    NegativeClassTopic^m = z^{m-} = \argmax_{z \in Z} \{ \sum_{i \in I_{test}} - \hat{y}_{i}^{m}\cdot \theta_z^i \}
\end{equation}
where: 
\begin{equation*}
    \hat{y}_{i}^{m} = 
    \begin{cases}
        1, & \hat{p}(y_{i}^{m}=1) \geq 0.5\\
        -1, & \hat{p}(y_{i}^{m}=1) < 0.5 
    \end{cases}
\end{equation*}
and $\hat{p}(y_{i}^{m}=1)$ is the positive class probability according to the DNN.\footnote{For our DNNs to produce class probabilities, we use a binary Cross-Entropy loss function~\cite{goodfellow2016deeplearning}, see Section~\ref{subsec:DNN_hyper}.}
We then investigate the $z^{m+}$ topic, which we denote as the \textit{positive class topic}, to observe what our model has learned.
From now on we shall refer to \textbf{$\hat{p}(y_{i}^{m}=1)$} as the \textbf{DNN's positive class prediction confidence} and to \textbf{$\theta_{z^{m+}}^i$} as the \textbf{positive class topic probability}.

Following~\cite{sievert2014ldavis},\footnote{Note that this work addresses topic model visualization, rather than neural network interpretation.} we print out the words which are most associated with the $z^{m+}$ topic according to their estimated term frequency within the selected topic. The top words belonging to those topics, for the game-level BERT-A-T classifiers, are presented in Table~\ref{tab:top-topics}. 

\begin{table}[htbp]
    \centering
    \scalebox{0.925}{
    \begin{tabular}{|l|l|l|l|l|l|l|} \hline
        PF & PTS & FGR & PR & SR & MSD2 & MSD3 \\ \hline
        Aggressive & Fun & Year & Try & Shot & Get & Game \\
        Defensively & Star & Team & Offensively & Miss & Shot & Last \\
        Guard & See & Season & Physical & Make & Ball & Night \\
        Adjustment & Great & Championship & Intensity & Three & Make & Make \\
        Transition & Think & Think & Pace & Field & Defense & Series \\
        Roll & Time & Career & Attack & Credit & Think & Win \\
        Turnover & Good & Work & Paint & Confident & Able & Little \\
        Paint & Enjoy & Great & Lane & Opportunity & Try & Think \\
        Defensive & Fan & Win & Communication & Look & Run & Better \\
        Offense & Basketball & Experience & Space & Ball & Go & Loss \\ \hline

    \end{tabular}}
    \caption{10 most likely words according to the topic with the largest probability difference between positive and negative BERT-A-T predictions, $z^{m+}$. We refer to these as the \textit{positive class topics}. Results are presented at the game-level, for each prediction task.}
    \label{tab:top-topics}
\end{table}

As can be seen in the table, the positive class topics seem intuitively related to the label, capturing not merely naive sentiment or emotion, but more refined, task related words. For positive personal fouls (PF) predictions, the model picks up concepts related to aggressive and defensive play. The topic for points (PTS) suggests it is most positively correlated with joyfulness, implying that the model has learned to associate a positive sentiment with more points. The field goal ratio (FGR) topic conveys game importance and long term thinking, implying a learned link between a player's performance and the game's significance. Pass risk (PR) and shot risk (SR) positive topics show a connection to players willingness to take more risk, as they are associated with words such as "try" and "confident", respectively. Finally, the mean shot distances metrics, MSD2 and MSD3, seem unrelated to their topics, which is unsurprising at least for MSD3 that BERT-A-T does not predict well (see results in Table~\ref{tab:results}).

\begin{figure}[htbp]
\includegraphics[width=\textwidth]{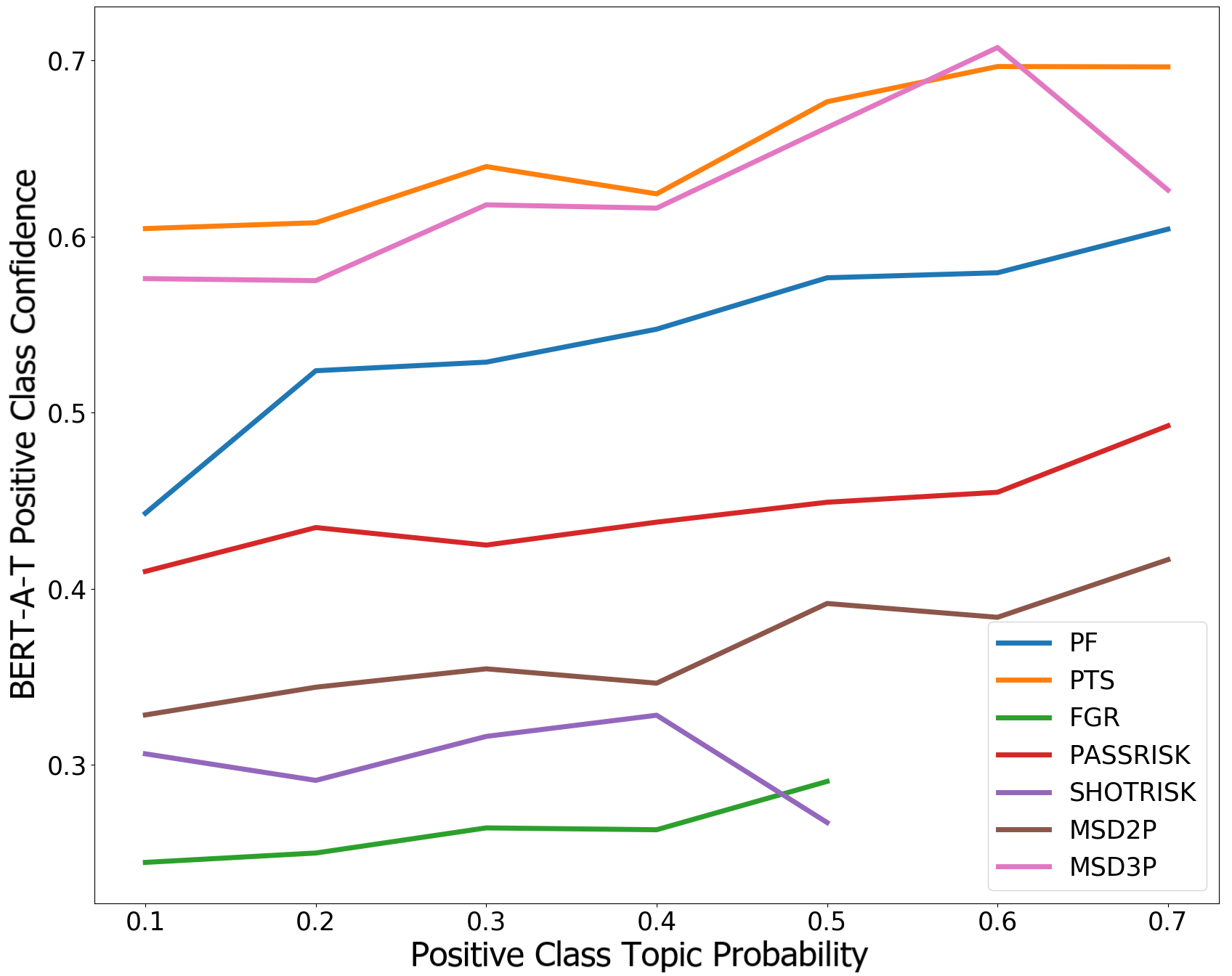}
\caption{BERT-A-T's averaged positive class prediction confidence ($\hat{p}(y_{i}^{m}=1)$) as a function of the positive class topic probability ($\theta_{z^{m+}}^i$) assigned to each interview. The computation of $f(\theta_{z^{m+}}^i; j)$, is described by Equation~\ref{eq:topics_preds}.}
\label{fig:topics_preds}
\end{figure}
In Figure \ref{fig:topics_preds} we plot $f(\theta_{z^{m+}}^i; j)$, the averaged BERT-A-T positive prediction confidence $\hat{p}(y_{i}^{m}=1)$, as a function of $\theta_{z^{m+}}^i$, the probability of the positive class topic $z^{m+}$ in the interviews, for each metric $m$. We compute each point in the plot, by calculating $f(\theta_{z^{m+}}^i; j)$ for each interval $j \in J=\{0.1, 0.2,..,1\}$ as defined:

\begin{equation}
    f(\theta_{z^{m+}}^i; j) = \frac{\sum_{i \in I^j_{test}} \hat{p}(y^{m}_{i}=1)}{|I^j_{test}|},\  I^j_{test} = \{ i\ |\ i \in I_{test}\ ,\ \theta_{z^{m+}}^i \in (j-0.1, j] \}
    \label{eq:topics_preds}
\end{equation}

We calculate this for visualization purposes: for each positive class topic probability ($\theta_{z^{m+}}^i$) interval of size $0.1$, we compute the average BERT-A-T positive prediction confidence ($\hat{p}(y_{i}^{m}=1)$) over all interviews with $\theta_{z^{m+}}^i$ falling in the interval.
As demonstrated in Figure~\ref{fig:topics_preds}, we observe that the probability assigned to positive topics ($\theta_{z^{m+}}^i$) from Table~\ref{tab:top-topics} increases monotonously with BERT-A-T's averaged positive predicted probability (confidence) for deviation above the player's mean performance ($\hat{p}(y_{i}^{m}=1)$), in $6$ out of the $7$ tasks (SR being the exception). 

\subsection{Topics Correlation with LSTM-T and BERT-A-T}

We finally investigate the predictions of our two best-performing text-based models, LSTM-T and BERT-A-T, in terms of the LDA topics. Since LDA is trained on free text and does not observe the predictions of any DNN model, its topics can be used to analyze two different DNNs: LSTM-T and BERT-A-T in this case.

\begin{figure}[htbp]
\centering
\begin{subfigure}[b]{1.0\textwidth}
   \includegraphics[width=1.15\textwidth]{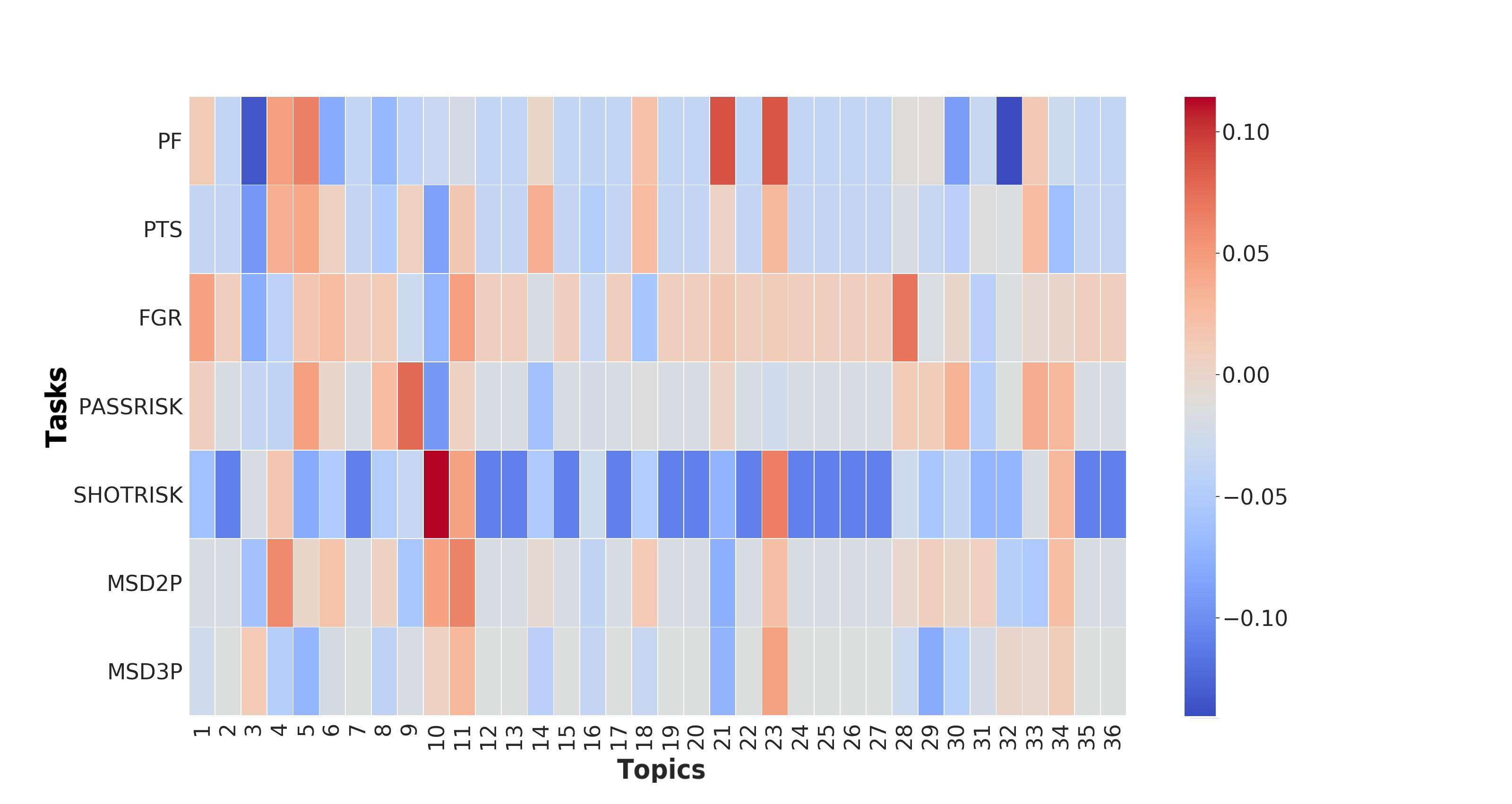}
   \caption{}
\end{subfigure}
\begin{subfigure}[b]{1.0\textwidth}
   \includegraphics[width=1.15\textwidth]{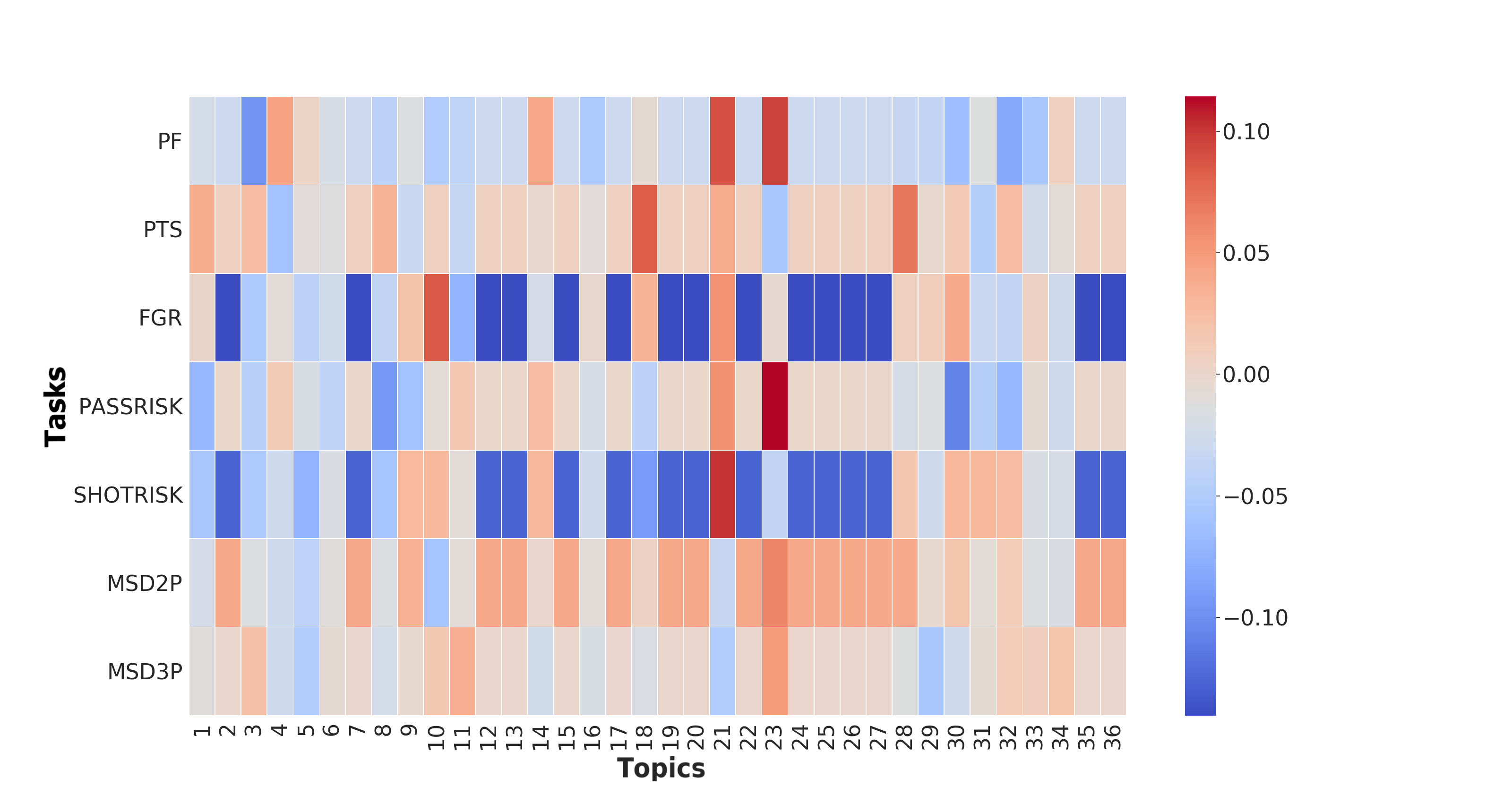}
   \caption{}
\end{subfigure}

\caption{Correlation heat-maps of LDA topic probability ($\theta_{z^m}^i$) and the DNN positive prediction confidence ($\hat{p}(y_{i}^{m}=1)$) (top: (a) LSTM-T; bottom: (b) BERT-A-T), for each prediction task.}
\label{fig:heat_map}
\end{figure}

The heat-maps in Figure~\ref{fig:heat_map} present the correlations for each metric $m$, between $\theta_{z^m}^i$, the probability of each topic $z \in Z$, and $\hat{p}(y_{i}^{m}=1)$, the DNN's positive prediction confidence, for both LSTM-T and BERT-A-T. Observing the results, it is clear that BERT-A-T's predictions are more correlated with the LDA topics, compared to the LSTM-T (darker colors indicate higher correlations). Looking closer at the heat-maps, some interesting thoughts come to mind. 

First, SR and MSD2 seem to exhibit opposing topic correlation patterns when examining the BERT-A-T model's predictions (bottom figure). This implies that topics associated with a higher probability of taking more 3 point shots are also associated with 2 point shots that are closer to the rim. The correlation exhibited by the BERT-A-T model could be explained by the growing tendency of players to take more 3 point shots at the expense of long-range 2 point shots, a phenomenon widely observed in recent years in the NBA.\footnote{See the substitutionality of long-range 2 point shots and 3 point shots in: \url{http://www.nbaminer.com/shot-distances/}} This phenomenon also serves as an example of attempting to maximize the expected payoff~\cite{von1944theory}. If shot-takers are in a situation in which they are unable to minimize their risk, i.e. they can not take a 2 point shot close to the rim and must take a shot further out, they choose to increase their risk for a higher payoff and take a longer-range 3 point shot.

Second, we also note that topic $23$, which is associated with PR for BERT-A-T, is also positively correlated with PF for both BERT-A-T and LSTM-T. PR's positive topic exhibits top-words such as "Physical" and "Intensity" (as can be seen in Table~\ref{tab:top-topics}), and PF's positive topic exhibits top-words such as "Aggressive", "Defensively" and "Guard". This suggests that topic $23$ captures some aspects of the concept of physical intensity and aggressiveness on both offense (PR) and defense (PF), and moreover that the DNNs capture this as well, to a certain extent.

Finally, the BERT-A-T classifiers for SR and FGR are generally correlated with similar topics. Particularly, it seems that both classifiers are positively correlated with topics that are highly associated with shot-related words.\footnote{We chose not to present all topics and their corresponding top-words as most are not informative and do not add information that helps us answer our research questions.} While this may seem counter-intuitive, notice that the corresponding LSTM-T classifiers do not show this correlation and they also perform significantly worse on those tasks (as seen in Table~\ref{tab:results}).
Overall, we can see that our topic analysis enables a level of intuitive reasoning, which serves as a new tool for providing insights on both the data and the properties of the prediction models (DNNs).

\section{Conclusions}

We explored the task of predicting the relative performance of NBA players based on their pre-game interviews, building on the extensive computational work on performance prediction in sports analytics and on text classification in NLP. We hypothesized that these interviews hold valuable information regarding the players' current state and that this information could shed light on their in-game actions and performance. 

To facilitate such a study, we collected a dataset consisting of pre and post game interviews alongside in-game performance metrics from the game following the interviews, for 36 prominent NBA players. Most of the games in our dataset are part of a playoff series where the stakes are typically higher, player and team performance or outcomes are harder to predict, and key players are expected to take on bigger roles, in comparison to regular season games.

Based on standard basketball in-game performance metrics, we introduced $7$ decision related metrics, each aiming to examine a different aspect of a player's in-game actions and performance. We formulated binary text classification tasks, attempting to predict a player's deviation from his mean performance in each metric, to examine whether pre and post game interview texts hold predictive power of the player's in-game actions. We formulated $4$ research questions, trying to assert: (1) if text classification models could utilize pre-game interviews for predicting performance, (2) if text could be combined with past performance metrics for better predictions, (3) if text classification models gain from modeling an interview's textual structure and (4) if DNNs could jointly learn a textual representation together with the task classifier, in a way that improves prediction accuracy over standard bag-of-words features.

We demonstrated that deep neural networks, mainly BERT and to some extent LSTM, are capable of using interview-based signals to predict players performance indicators. Moreover, we showed that our text-based models perform better than commonly used autoregressive metrics-based models, and that models which combine the two signals yield even better predictions. Also, we have shown that all game-level metrics and most period-level metrics can be predicted, to varying degrees.

In conducting this research, we were not interested only in predictive power, but also in understanding the phenomena at hand. Hence, it was important for us to be able to understand the predictions of the model. However, DNN interpretation is challenging, which hurts our ability to understand our models' predictions in terms of difference in language usage.

Thus, in order to interpret our DNN models in terms of linguistic concepts, we presented an LDA-based method. Our idea is based on examining topics which are on average most associated with the predictions of the model. In this analysis, we found that in 6 of 7 cases our best performing text-only model, BERT-A-T, is most associated with topics that are intuitively related to each prediction task, revealing that the model has successfully learned task specific representations.

Particularly, in this work we have attempted to set the stage for future exploration on several axes. First, much more research can be done on prediction models in scenarios where the signal is not stated directly in the text, in contrast to better explored cases such as sentiment classification and intent detection. Moreover, the connection between language and decision making can also be further explored, specifically in real-world rather than lab conditions. Models for predicting behavior using language could be useful in strategic settings such as multiplayer games, and can also be useful in medical settings, where predicting the patient's current state, which may not be explicitly discussed in the text, could have life-saving implications.

We consider our task a domain specific instance of a more general question: how should language be used as a predictive signal for actions in real world scenarios. We hope that our discussion, observations and algorithms, as well as dataset and code, will facilitate future work by the community. 

In future work we will consider more advanced time-series analysis tools along with the text, and better incorporate the interview's structure into our model. Our experiments have shown that incorporating the Q-A structure of the interview does improve the predictions, but this is only a first step in this direction, and more work can be done on this frontier. Also, we will aim to learn various performance metrics jointly, and to model the interactions between the actions of different players. This will hopefully yield better, more interpretable models for understanding the connection between language and actions. 

Another interesting direction is taking into account the development of players' performance over time. It could be that measuring players across fewer seasons could lead to better estimates, as players change along their careers. On average players are represented in our dataset for six seasons, which corresponds to their prime in terms of average performance. That is, we are generally capturing players at their peak performance, where quality tends to remain more consistent. In the future, given sufficient data per player, it would be interesting to integrate the time factor into our models.

Finally, another interesting theoretical question that arises is the extent to which language can really tell us about decisions that are made in later situations. It would be interesting to investigate and model these connections, and perhaps develop a theoretical bound for the predictive power that language could have in such situations.

\newpage
\bibliographystyle{compling}
\starttwocolumn
\bibliography{compling}

\end{document}